%% file: egpaper_for_review.tex
\documentclass[10pt,twocolumn,letterpaper]{article}

\usepackage{cvpr}
\usepackage{times}
\usepackage{epsfig}
\usepackage{graphicx}
\usepackage{amsmath}
\usepackage{amssymb}

\usepackage{relsize}
\usepackage{siunitx}
\usepackage{array,multirow,adjustbox}
\usepackage{pifont}
\usepackage{color}

\newcommand{\widthscalefive}{0.145}

\definecolor{orange}{rgb}{1,0.5,0}

\newcommand{\cmark}{\text{\ding{51}}}

\usepackage[pagebackref=true,breaklinks=true,letterpaper=true,colorlinks,bookmarks=false]{hyperref}

\usepackage{tabularx}
\newcolumntype{Y}{>{\centering\arraybackslash}X}

\cvprfinalcopy % *** Uncomment this line for the final submission

\def\cvprPaperID{****} % *** Enter the CVPR Paper ID here

\ifcvprfinal\fi
\begin{document}

%%%%%%%%% TITLE
%\title{Exploring Cross-scale Non-local Attention for Recurrent Image Super-Resolution Network}
\title{Image Super-Resolution with Cross-Scale Non-Local Attention \\ and Exhaustive Self-Exemplars Mining}

\author{Yiqun Mei$^{1}$, Yuchen Fan$^{1}$, Yuqian Zhou$^{1}$, Lichao Huang$^{2}$, \\
Thomas S. Huang$^{1}$, Humphrey Shi$^{3,1}$\\
\\
{\small $^1$IFP Group, UIUC, $^2$Horizon Robotics, $^3$University of Oregon}}

% \and

\maketitle
\thispagestyle{empty}

%%%%%%%%% ABSTRACT
\input{0_abstract.tex}
%%%%%%%%% INTRODUCTION
\input{1_Introduction.tex}
%%%%%%%%% RELATED WORK
\input{2_Related_work.tex}

%%%%%%%%% Introduction to CSNL idea
\input{3_attention.tex}

%%%%%%%%% Module Introduction
\input{4_Methodology.tex}

\input{5_Experiments.tex}

%%%%%%%%% CONCLUSION
\input{6_Conclusion.tex}

\paragraph{Acknowledgments}
This work is in part supported by IBM-Illinois Center for Cognitive Computing Systems Research (C3SR) - a research collaboration as part of the IBM AI Horizons Network.

{\small
\bibliographystyle{ieee_fullname}
\bibliography{egbib}
}

\end{document}

%%%%%%%%% TITLE
%\title{Exploring Cross-scale Non-local Attention for Recurrent Image Super-Resolution Network}
\title{Supplementary File: \\ Image Super-Resolution with Cross-Scale Non-Local Attention \\ and Exhaustive Self-Exemplars Mining}

\author{Yiqun Mei$^{1}$, Yuchen Fan$^{1}$, Yuqian Zhou$^{1}$, Lichao Huang$^{2}$, \\
Thomas S. Huang$^{1}$, Honghui Shi$^{3,1}$\\
\\
{\small $^1$IFP Group, UIUC, $^2$Horizon Robotics, $^3$University of Oregon}}

\maketitle
%\thispagestyle{empty}

%%%%%%%%% ABSTRACT
%\begin{abstract}
%In the supplementary material, we present following items:
%
%    1. Quantitative results for scale factor $\times$ 8 
%    
%    2. Quantitative results for scale factor $\times$ 8. 
%    
%    3. More visual comparisons.
%\end{abstract}

\section{Comparison with Na\"ive Cross-Scale Non-Local (CS-NL) Attention}
In the non-local structure, features are summed and weighted by corresponding spatial attention.
Formally, in-scale non-local attention is
\begin{equation}
    Z_{i, j} = \sum_{g, h}
    \frac{\exp(\phi(X_{i, j}, \textcolor{red}{X_{g, h}}))}{\sum_{u, v}\exp(\phi(X_{i, j}, X_{u, v}))}
    \psi(\textcolor{blue}{X_{g, h}})
\end{equation}
where \textcolor{red}{red} and \textcolor{blue}{blue} are the same features representation.

Na\"ive cross-scale non-local attention can be straightforwardly evolved as
\begin{equation}
    Z_{i, j} = \sum_{g, h}
    \frac{\exp(\phi(X_{i, j}, \textcolor{red}{Y_{g, h}}))}{\sum_{u, v}\exp(\phi(X_{i, j}, Y_{u, v}))}
    \psi(\textcolor{blue}{Y_{g, h}})
\end{equation}
where \textcolor{red}{red} and \textcolor{blue}{blue} are still the same but changed to $Y= X\downarrow_s$, that are the down-scaled features by by scaling factor s. The na\"ive cross-scale attention is build upon the correlation between features in different scales but summarises down-scaled features. The down-scaling operation will eliminate high-frequency details and lead performance regression in super-resolution tasks. 

The proposed cross-scale non-local attention summaries corresponding features in target scale without down-scaling operation, and can be formalized as
\begin{equation}
Z_{si, sj}^{s \times s} = \sum_{g, h}
\frac{\exp{\phi(X_{i, j}, \textcolor{red}{Y_{g, h}})}}{\sum_{u, v}\exp{\phi(X_{i, j}, Y_{u, v})}}
\psi(\textcolor{blue}{X_{sg, sh}^{s \times s}}),
\label{eq:csa1}
\end{equation}
where \textcolor{red}{red} and \textcolor{blue}{blue} are in different scales but one-to-one corresponded spatially.
In this way, the proposed cross-scale attention can keep high-resolution information in feature maps, utilize the original self-exemplar hints and benefits super-resolution performance.

\begin{table}[h]
%\footnotesize
%\small
    \centering
    \begin{tabularx}{\linewidth}{|c|Y|Y|Y|}
    \hline
          & Proposed Cross-scale & Na\"ive Cross-scale & In-scale  \\
         \hline
         PSNR & \textbf{33.74} & 33.65 & 33.62 \\
         \hline
    \end{tabularx}
    \caption{Comparison with Na\"ive Cross-Scale Non-Local (CS-NL) Attention on Set14 \cite{zeyde2010single} ($\times$2).}
    \label{tab:naive_cs}
\end{table}

Experiments in Table~\ref{tab:naive_cs} shows that the na\"ive cross-scale attention is negligible better than in-scale one,
and the proposed cross-scale attention significantly outperforms other approaches.

\section{More Qualitative Comparison}
In Fig. \ref{tab:table 2}-\ref{tab:table 1}, we provide more visual results to compare with other state-of-the-art methods. One can see that our approach reconstructed better image details, demonstrating the superiority of the proposed CSNLN.

\begin{figure*}[htbp]
%\tiny
%\small
	\newlength\fsdttwofigBD
	\setlength{\fsdttwofigBD}{-5.0mm}
	\scriptsize
	\centering
	\begin{tabular}{cc}
	%\tiny
	%\scriptsize
	%\footnotesize
	%\small
		%\hspace{-0.4cm}
		\begin{adjustbox}{valign=t}
		%\tiny
			\begin{tabular}{c}
				\includegraphics[height=0.25\textwidth, width=0.229\textwidth]{latex/supplementary/u1/hr_draw.png}
				\\
				 Urban100 ($4\times$):
				\\
				img\_046
				%\textsc{B100}: img_092
				
			\end{tabular}
		\end{adjustbox}
		\hspace{-2.3mm}
		\begin{adjustbox}{valign=t}
		%\tiny
			\begin{tabular}{cccccc}
				\includegraphics[width=\widthscalefive \textwidth]{latex/supplementary/u1/HRcrop.png} \hspace{\fsdttwofigBD} &
				\includegraphics[width=\widthscalefive \textwidth]{latex/supplementary/u1/bicubiccrop.png} \hspace{\fsdttwofigBD} &
				\includegraphics[width=\widthscalefive \textwidth]{latex/supplementary/u1/lapsrncrop.png} \hspace{\fsdttwofigBD} &
				\includegraphics[width=\widthscalefive \textwidth]{latex/supplementary/u1/edsrcrop.png} \hspace{\fsdttwofigBD} &
				\includegraphics[width=\widthscalefive \textwidth]{latex/supplementary/u1/dbpncrop.png} 
				\\
				HR \hspace{\fsdttwofigBD} &
				Bicubic \hspace{\fsdttwofigBD} &
				LapSRN~\cite{lai2017deep} \hspace{\fsdttwofigBD} &
				EDSR~\cite{lim2017enhanced} \hspace{\fsdttwofigBD} &
				DBPN~\cite{haris2018deep}
				\\
				\includegraphics[width=\widthscalefive \textwidth]{latex/supplementary/u1/oisrcrop.png} \hspace{\fsdttwofigBD} &
				\includegraphics[width=\widthscalefive \textwidth]{latex/supplementary/u1/rdncrop.png} \hspace{\fsdttwofigBD} &
				\includegraphics[width=\widthscalefive \textwidth]{latex/supplementary/u1/rcancrop.png} \hspace{\fsdttwofigBD} &
				\includegraphics[width=\widthscalefive \textwidth]{latex/supplementary/u1/sancrop.png} \hspace{\fsdttwofigBD} &
				\includegraphics[width=\widthscalefive \textwidth]{latex/supplementary/u1/ourcrop.png}  
				\\ 
				OISR~\cite{he2019ode} \hspace{\fsdttwofigBD} &
				RDN~\cite{zhang2018residual} \hspace{\fsdttwofigBD} &
				RCAN~\cite{zhang2018image} \hspace{\fsdttwofigBD} &
				SAN~\cite{dai2019second}  \hspace{\fsdttwofigBD} &
				Ours
			 \hspace{\fsdttwofigBD} 
				\\
			\end{tabular}
		\end{adjustbox}
		\vspace{0.5mm}
		\\

		\begin{adjustbox}{valign=t}
		%\tiny
			\begin{tabular}{c}
				\includegraphics[width=0.229\textwidth, height =0.25\textwidth]{latex/supplementary/u2/hr_draw.png}
				\\
				 Urban100 ($4\times$):
				\\
				img\_091
				%\textsc{B100}: img_092
				
			\end{tabular}
		\end{adjustbox}
		\hspace{-2.3mm}
		\begin{adjustbox}{valign=t}
		%\tiny
			\begin{tabular}{cccccc}
				\includegraphics[width=\widthscalefive \textwidth]{latex/supplementary/u2/HRcrop.png} \hspace{\fsdttwofigBD} &
				\includegraphics[width=\widthscalefive \textwidth]{latex/supplementary/u2/bicubiccrop.png} \hspace{\fsdttwofigBD} &
				\includegraphics[width=\widthscalefive \textwidth]{latex/supplementary/u2/lapsrncrop.png} \hspace{\fsdttwofigBD} &
				\includegraphics[width=\widthscalefive \textwidth]{latex/supplementary/u2/edsrcrop.png} \hspace{\fsdttwofigBD} &
				\includegraphics[width=\widthscalefive \textwidth]{latex/supplementary/u2/dbpncrop.png} 
				\\
				HR \hspace{\fsdttwofigBD} &
				Bicubic \hspace{\fsdttwofigBD} &
				LapSRN~\cite{lai2017deep} \hspace{\fsdttwofigBD} &
				EDSR~\cite{lim2017enhanced} \hspace{\fsdttwofigBD} &
				DBPN~\cite{haris2018deep}
				\\
				\includegraphics[width=\widthscalefive \textwidth]{latex/supplementary/u2/oisrcrop.png} \hspace{\fsdttwofigBD} &
				\includegraphics[width=\widthscalefive \textwidth]{latex/supplementary/u2/rdncrop.png} \hspace{\fsdttwofigBD} &
				\includegraphics[width=\widthscalefive \textwidth]{latex/supplementary/u2/rcancrop.png} \hspace{\fsdttwofigBD} &
				\includegraphics[width=\widthscalefive \textwidth]{latex/supplementary/u2/sancrop.png} \hspace{\fsdttwofigBD} &
				\includegraphics[width=\widthscalefive \textwidth]{latex/supplementary/u2/ourcrop.png}  
				\\ 
				
			    OISR~\cite{he2019ode} \hspace{\fsdttwofigBD} &
				RDN~\cite{zhang2018residual} \hspace{\fsdttwofigBD} &
				RCAN~\cite{zhang2018image} \hspace{\fsdttwofigBD} &
				SAN~\cite{dai2019second}  \hspace{\fsdttwofigBD} &
				Ours
				\\
			\end{tabular}
		\end{adjustbox}
		\vspace{0.5mm}
		\\
		%\hspace{-0.4cm}		
		\begin{adjustbox}{valign=t}
		%\tiny
			\begin{tabular}{c}
				\includegraphics[width=0.229\textwidth, height=0.25\textwidth]{latex/supplementary/u3/hr_draw.png}
				\\
				Urban100 ($4\times$):
				\\
				%\textsc{Urban100}: img\_004
				img\_093
			\end{tabular}
		\end{adjustbox}
		\hspace{-2.3mm}
		\begin{adjustbox}{valign=t}
		%\tiny
			\begin{tabular}{cccccc}
				\includegraphics[width=\widthscalefive \textwidth]{latex/supplementary/u3/HRcrop.png} \hspace{\fsdttwofigBD} &
				\includegraphics[width=\widthscalefive \textwidth]{latex/supplementary/u3/bicubiccrop.png} \hspace{\fsdttwofigBD} &
				\includegraphics[width=\widthscalefive \textwidth]{latex/supplementary/u3/lapsrncrop.png} \hspace{\fsdttwofigBD} &
				\includegraphics[width=\widthscalefive \textwidth]{latex/supplementary/u3/edsrcrop.png} \hspace{\fsdttwofigBD} &
				\includegraphics[width=\widthscalefive \textwidth]{latex/supplementary/u3/dbpncrop.png} 
				\\
				HR \hspace{\fsdttwofigBD} &
				Bicubic \hspace{\fsdttwofigBD} &
				LapSRN~\cite{lai2017deep} \hspace{\fsdttwofigBD} &
				EDSR~\cite{lim2017enhanced} \hspace{\fsdttwofigBD} &
				DBPN~\cite{haris2018deep}
				\\
				\includegraphics[width=\widthscalefive \textwidth]{latex/supplementary/u3/oisrcrop.png} \hspace{\fsdttwofigBD} &
				\includegraphics[width=\widthscalefive \textwidth]{latex/supplementary/u3/rdncrop.png} \hspace{\fsdttwofigBD} &
				\includegraphics[width=\widthscalefive \textwidth]{latex/supplementary/u3/rcancrop.png} \hspace{\fsdttwofigBD} &
				\includegraphics[width=\widthscalefive \textwidth]{latex/supplementary/u3/sancrop.png} \hspace{\fsdttwofigBD} &
				\includegraphics[width=\widthscalefive \textwidth]{latex/supplementary/u3/ourcrop.png}  
				\\ 
				OISR~\cite{he2019ode} \hspace{\fsdttwofigBD} &
				RDN~\cite{zhang2018residual} \hspace{\fsdttwofigBD} &
				RCAN~\cite{zhang2018image} \hspace{\fsdttwofigBD} &
				SAN~\cite{dai2019second}  \hspace{\fsdttwofigBD} &
				Ours
				\\
				
				\\
			\end{tabular}
		\end{adjustbox}
		\vspace{0.5mm}
		\\
		\begin{adjustbox}{valign=t}
		%\tiny
			\begin{tabular}{c}
				\includegraphics[width=0.229\textwidth, height=0.25\textwidth]{latex/supplementary/u4/hr_draw.png}
				\\
				Urban100 ($4\times$):
				\\
				%\textsc{Urban100}: img\_004
				img\_098
			\end{tabular}
		\end{adjustbox}
		\hspace{-2.3mm}
		\begin{adjustbox}{valign=t}
		%\tiny
			\begin{tabular}{cccccc}
				\includegraphics[width=\widthscalefive \textwidth]{latex/supplementary/u4/HRcrop.png} \hspace{\fsdttwofigBD} &
				\includegraphics[width=\widthscalefive \textwidth]{latex/supplementary/u4/bicubiccrop.png} \hspace{\fsdttwofigBD} &
				\includegraphics[width=\widthscalefive \textwidth]{latex/supplementary/u4/lapsrncrop.png} \hspace{\fsdttwofigBD} &
				\includegraphics[width=\widthscalefive \textwidth]{latex/supplementary/u4/edsrcrop.png} \hspace{\fsdttwofigBD} &
				\includegraphics[width=\widthscalefive \textwidth]{latex/supplementary/u4/dbpncrop.png} 
				\\
				HR \hspace{\fsdttwofigBD} &
				Bicubic \hspace{\fsdttwofigBD} &
				LapSRN~\cite{lai2017deep} \hspace{\fsdttwofigBD} &
				EDSR~\cite{lim2017enhanced} \hspace{\fsdttwofigBD} &
				DBPN~\cite{haris2018deep}
				\\
				\includegraphics[width=\widthscalefive \textwidth]{latex/supplementary/u4/oisrcrop.png} \hspace{\fsdttwofigBD} &
				\includegraphics[width=\widthscalefive \textwidth]{latex/supplementary/u4/rdncrop.png} \hspace{\fsdttwofigBD} &
				\includegraphics[width=\widthscalefive \textwidth]{latex/supplementary/u4/rcancrop.png} \hspace{\fsdttwofigBD} &
				\includegraphics[width=\widthscalefive \textwidth]{latex/supplementary/u4/sancrop.png} \hspace{\fsdttwofigBD} &
				\includegraphics[width=\widthscalefive \textwidth]{latex/supplementary/u4/ourcrop.png}  
				\\ 
				OISR~\cite{he2019ode} \hspace{\fsdttwofigBD} &
				RDN~\cite{zhang2018residual} \hspace{\fsdttwofigBD} &
				RCAN~\cite{zhang2018image} \hspace{\fsdttwofigBD} &
				SAN~\cite{dai2019second}  \hspace{\fsdttwofigBD} &
				Ours
				\\
				
				\\
			\end{tabular}
		\end{adjustbox}
	\end{tabular}
	\caption{
		Visual comparison for $4\times$ SR on Urban100 dataset.}
	\label{tab:table 2}
%\vspace{-3mm}
\end{figure*}

\begin{figure*}[htbp]
%\tiny
%\small
	\setlength{\fsdttwofigBD}{-5.0mm}
	\scriptsize
	\centering
	\begin{tabular}{cc}
	%\tiny
	%\scriptsize
	%\footnotesize
	%\small
		%\hspace{-0.4cm}
		\begin{adjustbox}{valign=t}
		%\tiny
			\begin{tabular}{c}
				\includegraphics[height=0.25\textwidth, width=0.229\textwidth]{latex/supplementary/1/hr_draw.png}
				\\
				 Manga109 ($4\times$):
				\\
				GakuenNoise
				%\textsc{B100}: img_092
				
			\end{tabular}
		\end{adjustbox}
		\hspace{-2.3mm}
		\begin{adjustbox}{valign=t}
		%\tiny
			\begin{tabular}{cccccc}
				\includegraphics[width=\widthscalefive \textwidth]{latex/supplementary/1/HRcrop.png} \hspace{\fsdttwofigBD} &
				\includegraphics[width=\widthscalefive \textwidth]{latex/supplementary/1/bicubiccrop.png} \hspace{\fsdttwofigBD} &
				\includegraphics[width=\widthscalefive \textwidth]{latex/supplementary/1/srcnncrop.png} \hspace{\fsdttwofigBD} &
				\includegraphics[width=\widthscalefive \textwidth]{latex/supplementary/1/vdsrcrop.png} \hspace{\fsdttwofigBD} &
				\includegraphics[width=\widthscalefive \textwidth]{latex/supplementary/1/lapsrncrop.png} 
				\\
				HR \hspace{\fsdttwofigBD} &
				Bicubic \hspace{\fsdttwofigBD} &
				SRCNN~\cite{dong2014learning} \hspace{\fsdttwofigBD} &
				VDSR~\cite{kim2016accurate} \hspace{\fsdttwofigBD} &
				LapSRN~\cite{lai2017deep}
				\\
				\includegraphics[width=\widthscalefive \textwidth]{latex/supplementary/1/drrncrop.png} \hspace{\fsdttwofigBD} &
				\includegraphics[width=\widthscalefive \textwidth]{latex/supplementary/1/edsrcrop.png} \hspace{\fsdttwofigBD} &
				\includegraphics[width=\widthscalefive \textwidth]{latex/supplementary/1/dbpncrop.png} \hspace{\fsdttwofigBD} &
				\includegraphics[width=\widthscalefive \textwidth]{latex/supplementary/1/rdncrop.png} \hspace{\fsdttwofigBD} &
				\includegraphics[width=\widthscalefive \textwidth]{latex/supplementary/1/ourcrop.png}  
				\\ 
				DRRN~\cite{tai2017image} \hspace{\fsdttwofigBD} &
				EDSR~\cite{lim2017enhanced} \hspace{\fsdttwofigBD} &
				DBPN~\cite{haris2018deep} \hspace{\fsdttwofigBD} &
				RDN~\cite{zhang2018residual}  \hspace{\fsdttwofigBD} &
				Ours 
			 \hspace{\fsdttwofigBD} 
				\\
			\end{tabular}
		\end{adjustbox}
		\vspace{0.5mm}
		\\

		\begin{adjustbox}{valign=t}
		%\tiny
			\begin{tabular}{c}
				\includegraphics[width=0.229\textwidth, height =0.25\textwidth]{latex/supplementary/2/hr_draw.png}
				\\
				 Manga109 ($4\times$):
				\\
				EverydayOsakanaChan
				%\textsc{B100}: img_092
				
			\end{tabular}
		\end{adjustbox}
		\hspace{-2.3mm}
		\begin{adjustbox}{valign=t}
		%\tiny
			\begin{tabular}{cccccc}
				\includegraphics[width=\widthscalefive \textwidth]{latex/supplementary/2/HRcrop.png} \hspace{\fsdttwofigBD} &
				\includegraphics[width=\widthscalefive \textwidth]{latex/supplementary/2/bicubiccrop.png} \hspace{\fsdttwofigBD} &
				\includegraphics[width=\widthscalefive \textwidth]{latex/supplementary/2/srcnncrop.png} \hspace{\fsdttwofigBD} &
				\includegraphics[width=\widthscalefive \textwidth]{latex/supplementary/2/vdsrcrop.png} \hspace{\fsdttwofigBD} &
				\includegraphics[width=\widthscalefive \textwidth]{latex/supplementary/2/lapsrncrop.png} 
				\\
				HR \hspace{\fsdttwofigBD} &
				Bicubic \hspace{\fsdttwofigBD} &
				SRCNN~\cite{dong2014learning} \hspace{\fsdttwofigBD} &
				VDSR~\cite{kim2016accurate} \hspace{\fsdttwofigBD} &
				LapSRN~\cite{lai2017deep}
				\\
				\includegraphics[width=\widthscalefive \textwidth]{latex/supplementary/2/drrncrop.png} \hspace{\fsdttwofigBD} &
				\includegraphics[width=\widthscalefive \textwidth]{latex/supplementary/2/edsrcrop.png} \hspace{\fsdttwofigBD} &
				\includegraphics[width=\widthscalefive \textwidth]{latex/supplementary/2/dbpncrop.png} \hspace{\fsdttwofigBD} &
				\includegraphics[width=\widthscalefive \textwidth]{latex/supplementary/2/rdncrop.png} \hspace{\fsdttwofigBD} &
				\includegraphics[width=\widthscalefive \textwidth]{latex/supplementary/2/ourcrop.png}  
				\\ 
				DRRN~\cite{tai2017image} \hspace{\fsdttwofigBD} &
				EDSR~\cite{lim2017enhanced} \hspace{\fsdttwofigBD} &
				DBPN~\cite{haris2018deep} \hspace{\fsdttwofigBD} &
				RDN~\cite{zhang2018residual}  \hspace{\fsdttwofigBD} &
				Ours 
			 \hspace{\fsdttwofigBD} 
				\\
			\end{tabular}
		\end{adjustbox}
		\vspace{0.5mm}
		\\
		%\hspace{-0.4cm}		
		\begin{adjustbox}{valign=t}
		%\tiny
			\begin{tabular}{c}
				\includegraphics[width=0.229\textwidth, height=0.25\textwidth]{latex/supplementary/3/hr_draw.png}
				\\
				Manga109 ($4\times$):
				\\
				%\textsc{Urban100}: img\_004
				Hamlet
			\end{tabular}
		\end{adjustbox}
		\hspace{-2.3mm}
		\begin{adjustbox}{valign=t}
		%\tiny
			\begin{tabular}{cccccc}
				\includegraphics[width=\widthscalefive \textwidth]{latex/supplementary/3/HRcrop.png} \hspace{\fsdttwofigBD} &
				\includegraphics[width=\widthscalefive \textwidth]{latex/supplementary/3/bicubiccrop.png} \hspace{\fsdttwofigBD} &
				\includegraphics[width=\widthscalefive \textwidth]{latex/supplementary/3/srcnncrop.png} \hspace{\fsdttwofigBD} &
				\includegraphics[width=\widthscalefive \textwidth]{latex/supplementary/3/vdsrcrop.png} \hspace{\fsdttwofigBD} &
				\includegraphics[width=\widthscalefive \textwidth]{latex/supplementary/3/lapsrncrop.png} 
				\\
				HR \hspace{\fsdttwofigBD} &
				Bicubic \hspace{\fsdttwofigBD} &
				SRCNN~\cite{dong2014learning} \hspace{\fsdttwofigBD} &
				VDSR~\cite{kim2016accurate} \hspace{\fsdttwofigBD} &
				LapSRN~\cite{lai2017deep}
				\\
				\includegraphics[width=\widthscalefive \textwidth]{latex/supplementary/3/drrncrop.png} \hspace{\fsdttwofigBD} &
				\includegraphics[width=\widthscalefive \textwidth]{latex/supplementary/3/edsrcrop.png} \hspace{\fsdttwofigBD} &
				\includegraphics[width=\widthscalefive \textwidth]{latex/supplementary/3/dbpncrop.png} \hspace{\fsdttwofigBD} &
				\includegraphics[width=\widthscalefive \textwidth]{latex/supplementary/3/rdncrop.png} \hspace{\fsdttwofigBD} &
				\includegraphics[width=\widthscalefive \textwidth]{latex/supplementary/3/ourcrop.png}  
				\\ 
				DRRN~\cite{tai2017image} \hspace{\fsdttwofigBD} &
				EDSR~\cite{lim2017enhanced} \hspace{\fsdttwofigBD} &
				DBPN~\cite{haris2018deep} \hspace{\fsdttwofigBD} &
				RDN~\cite{zhang2018residual}  \hspace{\fsdttwofigBD} &
				Ours
				\\
				
				\\
			\end{tabular}
		\end{adjustbox}
		\vspace{0.5mm}
		\\
		\begin{adjustbox}{valign=t}
		%\tiny
			\begin{tabular}{c}
				\includegraphics[width=0.229\textwidth, height=0.25\textwidth]{latex/supplementary/4/hr_draw.png}
				\\
				Manga109 ($4\times$):
				\\
				%\textsc{Urban100}: img\_004
				YumeiroCooking
			\end{tabular}
		\end{adjustbox}
		\hspace{-2.3mm}
		\begin{adjustbox}{valign=t}
		%\tiny
			\begin{tabular}{cccccc}
				\includegraphics[width=\widthscalefive \textwidth]{latex/supplementary/4/HRcrop.png} \hspace{\fsdttwofigBD} &
				\includegraphics[width=\widthscalefive \textwidth]{latex/supplementary/4/bicubiccrop.png} \hspace{\fsdttwofigBD} &
				\includegraphics[width=\widthscalefive \textwidth]{latex/supplementary/4/srcnncrop.png} \hspace{\fsdttwofigBD} &
				\includegraphics[width=\widthscalefive \textwidth]{latex/supplementary/4/vdsrcrop.png} \hspace{\fsdttwofigBD} &
				\includegraphics[width=\widthscalefive \textwidth]{latex/supplementary/4/lapsrncrop.png} 
				\\
				HR \hspace{\fsdttwofigBD} &
				Bicubic \hspace{\fsdttwofigBD} &
				SRCNN~\cite{dong2014learning} \hspace{\fsdttwofigBD} &
				VDSR~\cite{kim2016accurate} \hspace{\fsdttwofigBD} &
				LapSRN~\cite{lai2017deep}
				\\
				\includegraphics[width=\widthscalefive \textwidth]{latex/supplementary/4/drrncrop.png} \hspace{\fsdttwofigBD} &
				\includegraphics[width=\widthscalefive \textwidth]{latex/supplementary/4/edsrcrop.png} \hspace{\fsdttwofigBD} &
				\includegraphics[width=\widthscalefive \textwidth]{latex/supplementary/4/dbpncrop.png} \hspace{\fsdttwofigBD} &
				\includegraphics[width=\widthscalefive \textwidth]{latex/supplementary/4/rdncrop.png} \hspace{\fsdttwofigBD} &
				\includegraphics[width=\widthscalefive \textwidth]{latex/supplementary/4/ourcrop.png}  
				\\ 
				DRRN~\cite{tai2017image} \hspace{\fsdttwofigBD} &
				EDSR~\cite{lim2017enhanced} \hspace{\fsdttwofigBD} &
				DBPN~\cite{haris2018deep} \hspace{\fsdttwofigBD} &
				RDN~\cite{zhang2018residual}  \hspace{\fsdttwofigBD} &
				Ours
				\\
				
				\\
			\end{tabular}
		\end{adjustbox}
	\end{tabular}
	\caption{
		Visual comparison for $4\times$ SR on Manga109 dataset.}
	\label{tab:table 1}
%\vspace{-3mm}
\end{figure*}

{\small
\bibliographystyle{ieee_fullname}
\bibliography{egbib}
}

%% file: 0_abstract.tex
%%%%%%%%% ABSTRACT
\begin{abstract}
%Deep convolution-based Single Image Super Resolution (SISR) network embraces the benefits of large-scale external image resources for local recovery, while most works ignore the long-range feature-wise similarities existing in natural images. Recent work exploring non-local attention modules successfully leverages this intrinsic feature correlations. However, no previous deep model studies another internal property of images, \textit{cross-scale feature correlation}. In this paper, we propose the first Cross-Scale Non-Local (CS-NL) attention module integrated into a recurrent neural network. By combining the new CS-NL prior with local and in-scale non-local priors in a powerful recurrent fusion cell, it suggests that we can find more cross-scale feature correlations within a single Low Resolution (LR) image. The SISR performance is also significant improved by exhaustively integrating all the possible priors. Extensive experiments demonstrate the effectiveness of the CS-NL module by achieving the state-of-the-art performance on multiple SR benchmarks.
Deep convolution-based single image super-resolution (SISR) networks embrace the benefits of learning from large-scale external image resources for local recovery, yet most existing works have ignored the long-range feature-wise similarities in natural images. Some recent works have successfully leveraged this intrinsic feature correlation by exploring non-local attention modules. However, none of the current deep models have studied another inherent property of images: \textbf{cross-scale feature correlation}. In this paper, we propose the first Cross-Scale Non-Local (CS-NL) attention module with integration into a recurrent neural network. By combining the new CS-NL prior with local and in-scale non-local priors in a powerful recurrent fusion cell, we can find more cross-scale feature correlations within a single low-resolution (LR) image. The performance of SISR is significantly improved by exhaustively integrating all possible priors. Extensive experiments demonstrate the effectiveness of the proposed CS-NL module by setting new state-of-the-arts on multiple SISR benchmarks. Our code will be available at: \href{https://github.com/SHI-Labs/Cross-Scale-Non-Local-Attention}{https://github.com/SHI-Labs/Cross-Scale-Non-Local-Attention}
\end{abstract}

%% file: 1_Introduction.tex
\section{Introduction}
Single image super resolution (SISR) aims at recovering a high-resolution (HR) image from its low-resolution (LR) counterpart. SISR has numerous applications in the areas of satellite imaging, medical imaging, surveillance monitoring and high-definition display and imaging \textit{etc} \cite{demirel2011discrete, yu2017computed,zhou2018survey,zhou2020image,zou2011very}. The mapping between LR and HR image is not bijective, which yields more possibilities for a faithful and high-quality HR recovery. Due to this ill-posed nature, SISR remains challenging in the past decades. 
%Single image super resolution (SISR), which aims to reconstructing a high resolution (HR) image from its low resolution (LR) counterparts, is a challenging problem in computer vision and image processing, with numerous applications including satellite and medical imaging, surveillance monitoring and high-definite display. 
%One important reason that image SR remains challenging is due to its ill-posed nature, since there are many HR solutions corresponding to a same LR image. To alleviate this issue, early SR methods have investigated various image priors, either locally or non-locally and at the same or cross image scale, to regularize the solution space. 

Early efforts in traditional methods provide good practices for resolving SISR. By fully using the intrinsic property of the LR images, they mostly focus on local prior and non-local prior for patch matching and reconstruction. Specifically, local prior based methods, like bilinear or bicubic interpolation, reconstruct pixels merely by the weighted sum of neighbour ones. To go beyond the local limitation, methods based on non-local mean filtering \cite{protter2008generalizing, zhang2012single} start to globally search similar patches over the whole LR image. 

\begin{figure}[t]
    \centering
      \includegraphics[width=0.9\linewidth]{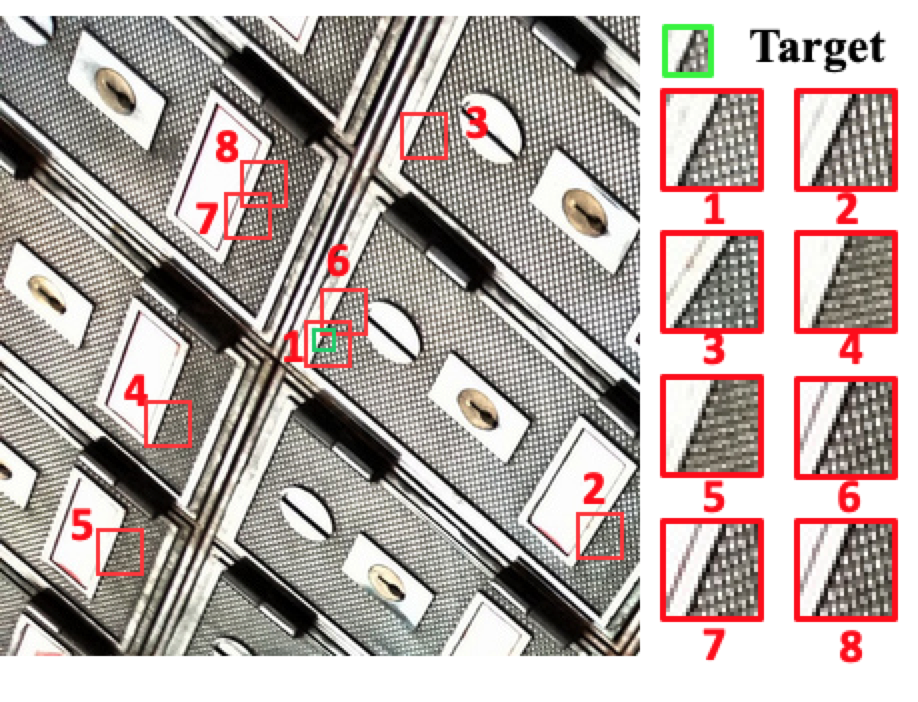}
    %\caption{Visualization of internal HR exemplars (red) given a target patch. We select top 8 matches generated by our cross-scale attention.}
    \caption{Visualization of most engaged patches captured by our Cross-Scale Non-Local (CS-NL) attention. Cross-scale similarities widely exist in natural images. Multiple high-resolution (HR) patches from the low-resolution (LR) image itself significantly improve target patch super-resolution.}
    \label{fig:match}
\end{figure}
The non-local search for self-similarity can be further extended to \textit{cross-scale} cues. It has been verified that cross-scale patch similarity widely exists in natural images \cite{glasner2009super,zontak2011internal}. Intuitively, in addition to non-local pixel-to-pixel matching, pixels can also be matched with larger image patches. The natural cross-scale feature correspondence makes us search high-frequency details directly from LR images, leading to more faithful, accurate and high-quality reconstructions.

Since the first deep learning-based method \cite{dong2014learning} was proposed, discriminative learning based methods make it possible to use large-scale external image priors for SISR. Compared with traditional methods, they tend to have better feature representation ability, faster inference speed, end-to-end trainable paradigm \cite{goodfellow2016deep,krizhevsky2012imagenet}, and significant performance improvement. To further take the advantages of deep SISR, for several years, efforts \cite{ fan2017balanced, kim2016accurate,lim2017enhanced,zhang2018image,zhang2018residual,yu2018wide,fan2019scale} have been made on increasing the depth or width of the networks to increase the receptive field or improve the feature representation. However, the essence of the solutions was not changed, but locally finding external similar patches. It yields great limitations of deep SISR. SISR performance was boosted right after the non-local attention modules  \cite{dai2019second,liu2018non,zhang2019residual} were proposed. They explored non-local self-similarity property and embedded the non-local modules into the deep network.

%Recently, deep convolution neural networks (CNNs) have become the mainstream for image SR, because of their powerful representation ability and end-to-end trainable paradigm  \cite{krizhevsky2012imagenet,he2016deep,lecun2015deep,goodfellow2016deep}. The first method utilized CNNs was proposed by Dong et al \cite{dong2014learning}, which achieved significant improvement over traditional SR approaches. In last several years, extensive researches have concentrated on increasing depth and width of networks to obtain better feature representation. Although considerable progress are obtained, their performance are limited by the small receptive fields, as they extracts features in a local way with convolution operations, which fails to capture long-range features dependencies. Recent state-of-art approach tackle this weakness by incorporating non-local attention modules \cite{zhang2019residual,dai2019second} into their networks to explicitly capture long-range feature correlation in a global fashion. 

What should be the next progress for deep SISR? One intuitive idea is following the traditional methods to explore the non-local \textit{cross-scale} self-similarity in deep networks. Recently, Shocher et al. \cite{shocher2018zero} proposed a zero-shot super-resolution (ZSSR) network to learn the high-frequency details from a pair of down-sampled LR and LR itself using one single test LR image. The essence of ZSSR is an implicit cross-scale patch matching approach using a light-weight network. However, inferring with ZSSR requires additional training time for each new LR image, which is not elegant and efficient enough for practical applications.

%Although tremendous efforts have brought unprecedented success for image SR, existing CNN based methods still face certain limitation: they fail to make use of the cross-scale self-similarities within the input images. The \textit{across scale} recurrence of a small piece of information was demonstrated to be a strong property of natural images. As shown in [cite], if well explored, the internal recurrence can serve as strong internal reference and bring stronger predictive power than external statistics to help the networks recover missing high frequency details, especially when such information is usually lost in the original scale or maybe even never existed in the training dataset. Recently, Shocher \cite{ZSSR} proposed ZSSR to reconstruct HR images solely rely on their internal information, by training a small network (ZSSR) at the test time to learn the mapping functions between LR image and its downscaled version. However, the slow running speed and limited capacity due to run-time training impedes it far from practical applications. Moreover, their performance may be potentially challenged by \textit{hard cases}, where internal correlation is insufficient to fully support accurate reconstruction. 

Following the successful path of non-local attention modules, in this paper, we are seeking ways of incorporating \textit{cross-scale} non-local attention scheme into the deep SR network. Specifically, we propose a novel Cross-Scale Non-Local (CS-NL) attention module, learning to mine long-range dependencies between LR features to larger-scale HR patches within the same feature map, as shown in Figure \ref{fig:match}. After that, we integrate the previous local prior, In-Scale Non-Local (IS-NL) prior and the proposed Cross-Scale Non-Local prior into a Self-Exemplars Mining (SEM) module, and fuse them with multi-branch mutual-projection. Finally, we embed the SEM module into a recurrent framework for image super-resolution task.

In summary, the main contributions of this paper are three-fold:
\begin{itemize}
    \item The core contribution of the paper is to propose the first Cross-Scale Non-Local (CS-NL) attention module in deep networks for SISR task. We explicitly formulate the pixel-to-patch and patch-to-patch similarities inside the image, and demonstrate that additionally mining cross-scale self-similarities greatly improves the SISR performance.
    \item We then propose a powerful Self-Exemplar Mining (SEM) cell to fuse information recurrently. Inside the cell, we exhaustively mine all the possible intrinsic priors by combining local, in-scale non-local, and the proposed cross-scale non-local feature correlations, and embrace rich external statistics learned by the network.
    \item The newly proposed recurrent SR network achieves the state-of-the-art performance on multiple image benchmarks. Extensive ablation experiments further verify the effectiveness of the novel network.
\end{itemize}

%% file: 2_Related_work.tex
%-------------------------------------------------------------------------
\section{Related Works}
%Since the development of image SR has a long history, a comprehensive review is beyond the scope of this paper. Here we focus on the most relevant and recent algorithms.

\paragraph{Self-Similarity in Image SR}
The fact that small patches tend to recur within and across scale of a same image has been verified for most natural images \cite{glasner2009super, zontak2011internal}. Since then, a category of self-similarity based approaches has been extensively developed and achieves promising results. Such algorithms utilize the cross-scale information redundancy of a given image as a unique source for reconstruction without relying on any external examples \cite{freedman2011image,freeman2002example,glasner2009super,huang2015single,michaeli2013nonparametric,singh2014super,yang2013fast}. In the pioneering work, Glasner \textit{et al.} \cite{glasner2009super} proposed to jointly exploit repeating patches within and across image scales by integrating the idea of multiple image SR and example-based SR into a unified framework. Furthermore, Freedman \textit{et al.} \cite{freedman2011image} effectively assumed that similar patches exist in an extremely localized region and thus can greatly reduce computation time. Following this fashion, Yang \textit{et al.} \cite{yang2013fast} proposed a very fast regression model that focused on only in-place cross-scale similarity. To handle appearance variations in the scene, Huang \textit{et al.} \cite{huang2015single} enlarged the internal dictionary by modeling geometric transformations. The idea of internal data repetition has also been applied to solve SR with blur and noisy images \cite{michaeli2013nonparametric,singh2014super}.

\paragraph{Deep CNNs for Image SR} 

The first work that introduced CNN to solve image SR was proposed by \cite{dong2014learning}, where they interpret the three consecutive convolution layers as corresponding extraction, non-linear mapping and reconstruction step in sparse coding. Kim \textit{et al.} \cite{kim2016accurate} proposed a very deep model VDSR with more than 16 convolution layers benefiting from effective residual learning. To further unleash the power of deep CNNs, Lim \textit{et al.} \cite{lim2017enhanced} integrated residual blocks into the SR framework to form a very wide model (EDSR) and a very deep model (MDSR). As the network goes as deep as hundreds of layers, Zhang \textit{et al.} \cite{zhang2018residual} utilized densely connected blocks with global feature fusion to effectively exploit hierarchical features from all intermediate layers. Besides extensive efforts spent on designing wider and deeper structures, algorithms with attention modules \cite{dai2019second,liu2018non,zhang2018image,zhang2019residual} were proposed to further enhance representation power of deep CNNs by exploring feature correlations along either spatial or channel dimension. 

\paragraph{Non-Local Attention in Deep Networks}
In recent years, there is an emerging trend of applying non-local attention mechanism to solve various computer vision problems. In general, non-local attention in deep CNNs allows the network to concentrate more on informative areas. Wang \textit{et al.} \cite{wang2018non} initially proposed non-local neural network to seek semantic relationships for high-level tasks, such as image classification and object detection. On the contrary, non-local attention for image restoration is based on non-local similarities prior. Methods, such as NRLN \cite{liu2018non}, RNAN \cite{zhang2018image} and SAN \cite{dai2019second}, incorporate non-local operation into their networks in order to make better use of image structural cues, by considering long-range feature correlations. As such, they achieved considerable performance gain.

However, existing non-local approaches for image restoration only explored feature similarities at the same scale, while ignoring abundant internal LR-HR exemplars across scales, leading to relatively low performance. It is known that the internal HR correspondences contain more relevant high-frequency information and stronger predictive power. To this end, we propose Cross-Scale Non-Local (CS-NL) attention by exploring cross-scale feature correlations.

%% file: 3_attention.tex
\section{Cross-Scale Non-Local (CS-NL) Attention}\label{sec:csnl}

In this section, we formulate the proposed cross-scale non-local attention, and compare it with the existing in-scale non-local attention. 
\paragraph{In-Scale Non-Local (IS-NL) Attention}
Non-local attention can explore self-exemplars by summarizing related features from the whole images.
Formally, given image feature map $X$, the non-local attention is defined as
\begin{equation}
    \label{eq:nla}
    Z_{i, j} = \sum_{g, h}
    \frac{\exp(\phi(X_{i, j}, X_{g, h}))}{\sum_{u, v}\exp(\phi(X_{i, j}, X_{u, v}))}
    \psi(X_{g, h}),
\end{equation}
where $(i, j)$, $(g, h)$ and $(u, v)$ are pairs of coordinates of $X$. $\psi(\cdot)$ is feature transformation function, and $\phi(\cdot,\cdot)$ is correlation function to measure similarity that is defined as
\begin{equation}
    \label{eq:phi}
    \phi(X_{i, j}, X_{g, h}) = \theta(X_{i,j})^T\delta(X_{g, h}),
\end{equation}
where $\theta(\cdot)$ and $\delta(\cdot)$ are feature transformations. Note that the pixel-wise correlation is measured in the same scale.

% Given a scale factor $s$ for Image SR and input feature map $X$ with size $H \times W \times C$, the goal of cross-scale matching is to synthesize a HR feature $Z$ with shape $sH \times sW \times C$ using internal HR hints in $X$. Inspired by traditional self-similarity SR, we define the cross-scale non-local operation as 
% \begin{align}
% Z_{sx,sy}^{s\times s} = \frac{\mathlarger{\sum}\limits_{\forall i,j} \phi\bigg(X_{x,y}, X_{i,j}^{s\times s}\bigg)}{\mathlarger{\sum}\limits_{\forall i,j} \phi\bigg(X_{x,y}, X_{i,j}^{s\times s}\bigg)} \psi(X_{i,j}^{s\times s})
% \end{align}
% where $X_{x,y}$ represents a pixel-wise feature (of shape $1\times 1 \times C$) at location $(x,y)$ in X and $X(Z)_{a,b}^{s\times s}$ is a region of features centered at $(a,b)$ with size $s \times s \times C$. The function $\phi$
% computes pairwise affinity between feature $X_{x,y}$ and region $X_{i,j}^{s\times s}$. The function $\psi$ computes a representation of the input region at position $(i,j)$.

\begin{figure}[t]
	\centering
		\includegraphics[clip, trim=0 2.2cm 0 2.2cm,width=\linewidth]{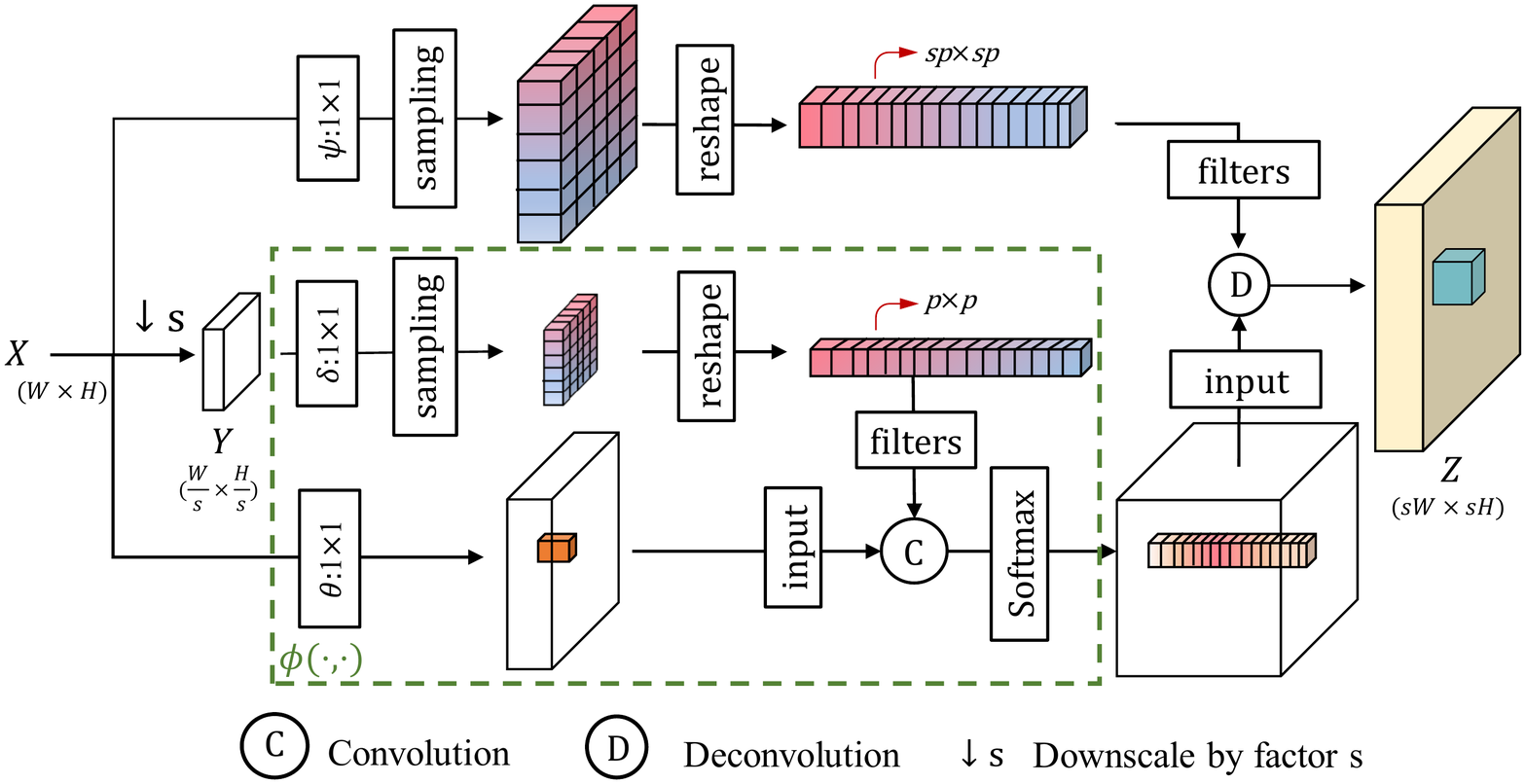}
	\caption{The proposed Cross-Scale Non-Local (CS-NL) attention module. The bottom green box is for patch-level cross-scale similarity-matching. The upper branch shows extracting the original HR patches in LR image. }
	\label{fig:attention}
\end{figure}

\paragraph{Cross-Scale Non-Local (CS-NL) Attention}

%Therefore, suppose the input feature map is X ($W \times H$), to compute pixel-patch similarity, we need to first down-sample $X$ to $Y$ ($\frac{W}{s} \times \frac{H}{s}$) and find pixel-wise similarity between $X$ and $Y$ (the $\phi$ function realized by $1 \times 1$ convolution), and finally use $s \times s$ patches in $X$ to super-resolve pixels in $X$, thus the output will be $sW \times sH$.

The above formulation can be easily extended to a cross-scale version referring to Figure \ref{fig:attention}. Instead of measuring the pixel-wise mutual correlation as the in-scale non-local module, the proposed cross-scale attention is designed to measure the correlation between low-resolution pixels and larger-scale patches in the LR image. To super-resolve the LR image, the Cross-Scale Non-Local (CS-NL) attention directly utilizes the patches matched to each pixel within this LR image.  

%In super-resolution tasks, we weighted average the matched patches from the  

%The proposed cross-scale attention is designed to weighted average similar high-resolution patches then super-resolve low-resolution pixels, according to correlation between $1\times1$ pixels and $s \times s$ patches in the features, instead of $1\times1$ pixels mutual correlation in previous non-local module.

%by measuring the cross-scale similarities between pixel-feature and larger-scale regions. In super-resolution tasks, the cross-scale non-local attention is designed to directly utilize similar patterns with richer details within the input image. 

Hence, for super-resolution purposes, cross-scale non-local attention is built upon in-scale attention by finding candidates in features $Y= X\downarrow_s$ downsampled by scaling factor $s$. The reason to do so is because directly matching pixels with patches using common similarity measurement is infeasible due to spatial dimension difference. So we simply downsample the features to represent the patch as pixel and measure the affinity. Downsampling operation in this paper is bilinear interpolation.

Suppose the input feature map is $X$ ($W \times H$), to compute pixel-patch similarity, we need to first downsample $X$ to $Y$ ($\frac{W}{s} \times \frac{H}{s}$) and find pixel-wise similarity between $X$ and $Y$, and finally use corresponding $s \times s$ patches in $X$ to super-resolve pixels in $X$, thus the output $Z$ will be $sW \times sH$.
%Suppose $X$ has a spatial size $(W, H)$, then $Y$, the matching candidate, is $(\frac{W}{s}, \frac{H}{s})$. The super-resolved output $Z$ will be $(sW, sH)$. 
Cross-scale attention can be adapted from Eq.\ref{eq:nla} as
\begin{equation}
Z_{si, sj}^{s \times s} = \sum_{g, h}
\frac{\exp({\phi(X_{i, j}, Y_{g, h})})}{\sum_{u, v}\exp{(\phi(X_{i, j}, Y_{u, v}))}}
\psi(X_{sg, sh}^{s \times s}),
\label{eq:csa1}
\end{equation}
where $Z_{si, sj}^{s \times s}$ now is the feature patch of size $s \times s$ located at $(si, sj)$. We obtain the weighted-averaged features $Z_{si, sj}^{s \times s}$ directly from the feature patches $X_{sg, sh}^{s \times s}$ extracted from the input feature maps. Intuitively, with the cross-scale attention, we can mine more faithful and richer high-frequency details from the original intrinsic image resources.

\begin{figure*}
   \centering
		\includegraphics[clip, trim=0cm 2cm 0cm 3cm, width=0.9\linewidth]{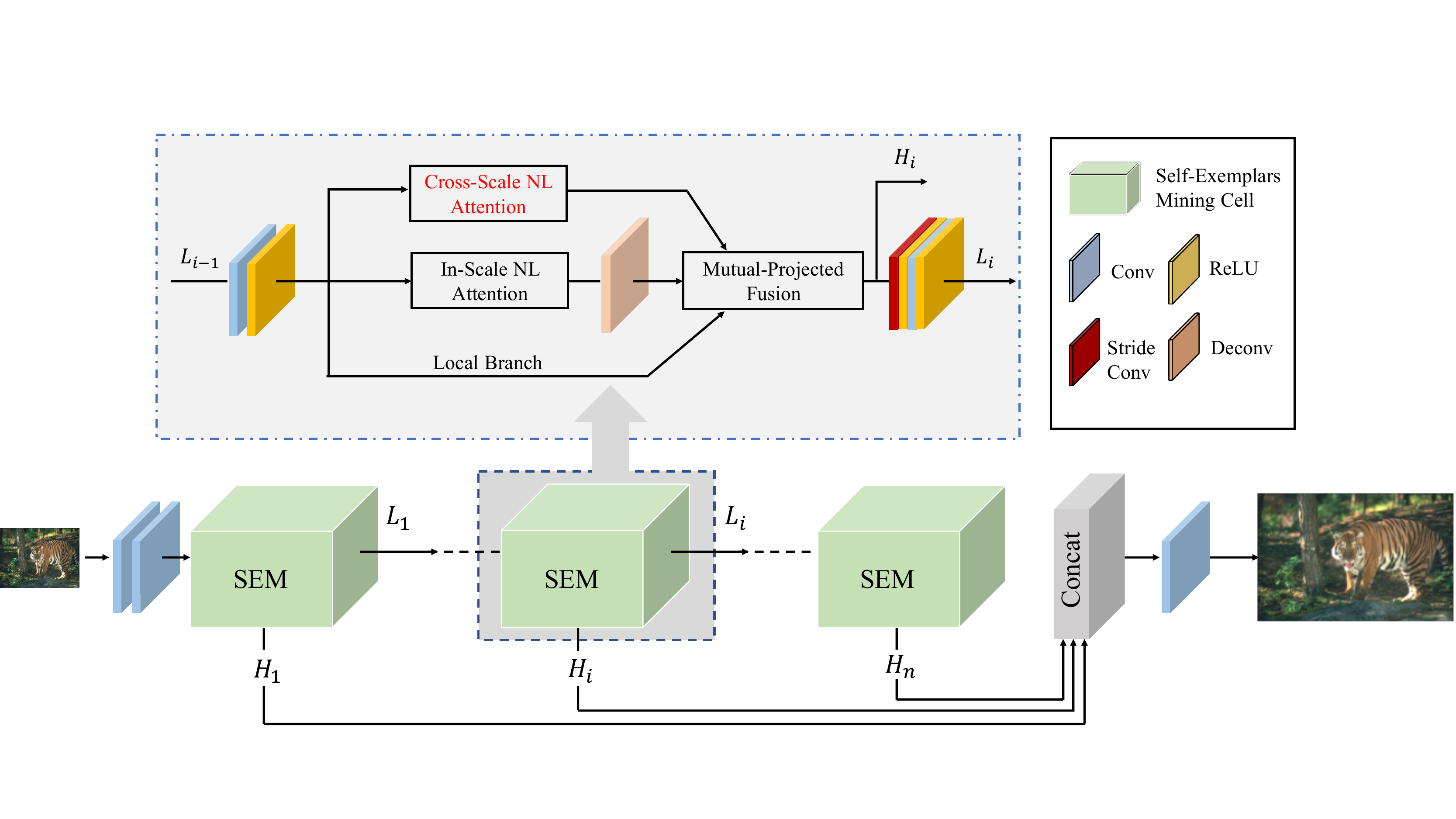}
		%\vspace{-6mm}
    \caption{The recurrent architecture with the proposed Self-Exemplars Mining (SEM) cell. Inside SEM, it fuses features learned from a proposed Cross-Scale Non-Local (CS-NL) attention, with others from the In-Scale Non-Local (IS-NL) and the local paths.} 
    \label{fig:model}
\end{figure*}

\paragraph{Patch-Based Cross-Scale Non-Local Attention}
Feature-wise affinity measurement can be problematic. First, high-level features are robust to transformations and distortions, that is rotated/distorted low-level patches may yield same high-level features. Take the average pooling as an example, an original region representing a HR \textit{window} and its flipped version have exactly the same high-level features. Therefore, it is likely that many erroneous matches will be synthesized to HR tensors. Besides, adjacent target regions (e.g. $Z_{si,sj}^{s\times s}$ and $Z_{s(i+1),s(j)}^{s\times s}$) are generated in a non-overlapping fashion, possibly creating discontinuous region boundaries artifacts.   
%Above general form provides a possible implementation to capture the internal hints by computing feature/feature-descriptor correlations. However, such 

Based on the above analysis, we generalize to empirically implement our Cross-Scale Non-Local (CS-NL) attention using another patch-wise matching. Therefore, Eq.\ref{eq:csa1} is generalized to,
\begin{equation}
    Z_{si, sj}^{sp \times sp} = \sum_{g, h}
    \frac{\exp{\phi(X_{i, j}^{p\times p}, Y_{g, h}^{p\times p})}}{\sum_{u, v}\exp{\phi(X_{i, j}^{p\times p}, Y_{u, v}^{p\times p})}}
    \psi(X_{sg, sh}^{sp \times sp}),
    \label{eq:pbcsa1}
\end{equation}
and Eq.\ref{eq:pbcsa1} will be identical to Eq.\ref{eq:csa1} if $p=1$. The measured correlations are efficiently extended to patch-level, and regions in the output feature map $Z$ are now densely overlapped due to patch-based matching. 
%As illustrated in Fig. \ref{fig:attnetion}, embedded descriptor $Y$ are densely sampled into $p\times p\times C$ patches with stride $s$. The correlations are efficiently evaluated at patch level through convolution operations over input feature map $X$, by grouping the cropped patches from embedded descriptor $Y$ into filters. After obtaining the matching score, we extract corresponding patches (in the sense of patches from $Y$) from $\psi(X)$ as deconvolution filters to synthesize the target SR feature. Regions in output feature map $Z$ are now densely overlapped due to the small-stride densely sampling strategy. Moreover, these overlapping areas are automatic averaged through deconvolution operation. 

%% file: 4_Methodology.tex
\section{Methodology}

The proposed network architecture is shown in Figure \ref{fig:model}. It is basically a recurrent neural network, with each recurrent cell called Self-Exemplars Mining (SEM) fully integrating local, in-scale non-local, and a newly proposed Cross-Scale Non-Local (CS-NL) priors. In this section, we introduce them in a bottom-up manner.

\input{./4_1_CS.tex}
\input{./4_2_fusion.tex}
\input{./4_3_framework.tex}

%% file: 4_1_CS.tex
\subsection{CS-NL Attention Module}

Figure \ref{fig:attention} illustrates the newly-proposed Cross-Scale Non-Local (CS-NL) attention module embedded into the deep networks. As formulated in section \ref{sec:csnl}, we apply a patch-level cross-scale similarity-matching in the CS-NL attention module. Specifically, suppose we are conducting an s-scale super-resolution with the module, given a feature map $X$ of spatial size $(W, H)$, we first bilinearly downsample it to $Y$ with scale $s$, and match the $p \times p$ patches in $X$ with the downsampled $p \times p$ candidates in $Y$ to obtain the softmax matching score. Finally, we conduct deconvolution on the score by weighted adding the patches of size $(sp, sp)$ extracted from $X$. The obtained $Z$ of size $(sW, sH)$, will be $s$ times super-resolved than $X$.

%% file: 4_2_fusion.tex
\subsection{Self-Exemplars Mining (SEM) Cell}

\paragraph{Multi-Branch Exemplars}
Inside the Self-Exemplars Mining (SEM) cell, we exhaustively mine all the possible intrinsic priors, and embrace rich external image priors. Specifically, we mine the image self-similarities and learn the new information using a multi-branch structure, including the conventional Local (L) and In-Scale Non-Local (IS-NL) branches, and also the newly proposed CS-NL branch.  

The local branch, in Figure \ref{fig:model}, is a simple identical pathway connecting the convolutional features to the fusion structure. For the IS-NL branch, it contains a non-local attention module adopted from \cite{dai2019second} and a deconvolution layer for upscaling the module outputs. The IS-NL module is region-based in this paper. As in \cite{dai2019second}, we divide the feature maps into region grids, where the inter-dependencies are captured independently in each grid. This reduces the computation burden.

\paragraph{Mutual-Projected Fusion}
\begin{figure}[t]
	\centering
		\includegraphics[width=0.9\linewidth]{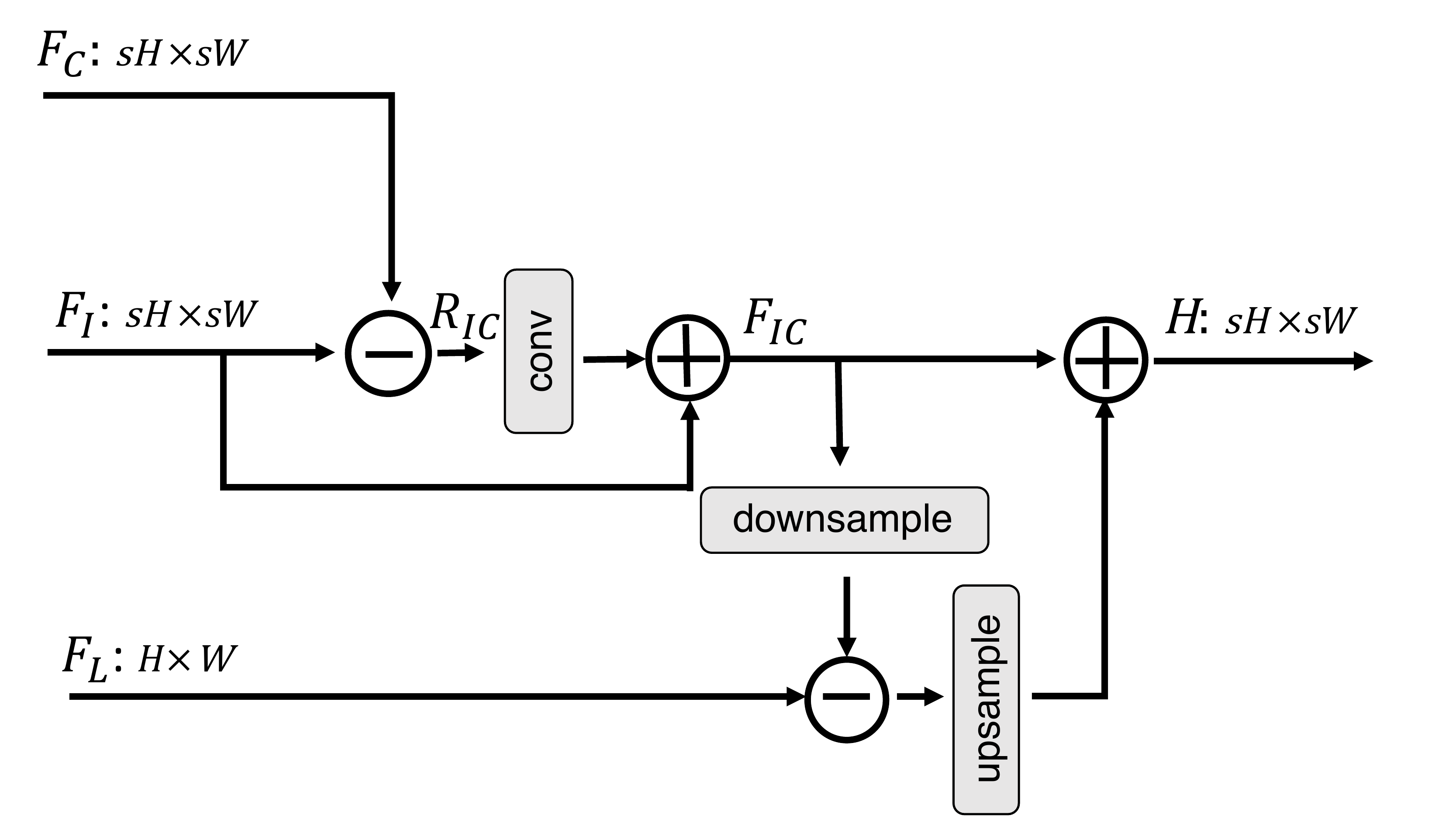}
	\caption{Mutual-projected fusion. Downsample and upsample operations are implemented using stride convolution and stride deconvolution, respectively.}
	\label{fig:fusion}
\end{figure}
While three-branch structure in SEM generates three feature maps by independently exploiting each information sources from LR images, how to fuse these separate tensors into a comprehensive feature map remains unclear. One possible solution is simply adding or concatenating them together, as widely used in previous works \cite{lim2017enhanced,liu2018non,zhang2019residual,zhang2018residual}. %However, we find that such operation is not optimal in our case, since feature from independent sources are less correlated. As a result, 
In this paper, we proposed a mutual-projected fusion to progressively combine features together. The algorithm procedure is illustrated in Figure \ref{fig:fusion}.

%\paragraph{xxx projection}
%Inspired by projection approach in recent video super resolution \qa{[cite]}, the proposed fusion method treats features from CS-NL, IS-NL and L branch as independent features, since they are computed from distinct sources. 

%In this approach, a residual tensor are iteratively calculated as reconstruction errors between a HR features and multiple corresponding features from other frame. The residuals are then projected back to the target feature to improve reconstruction. 

%In our case, it is sufficient to treat features reconstructed from non-local branch and cross-scale non-local branch as independent features, since they are synthesized from distinct sources. 

To allow the network to concentrate on more informative features, we first compute the residual $R_{IC}$ between two features from IS-NL $F_I$ and CS-NL $F_C$ branch, and then after a single convolution layer $conv$ on $R_{IC}$, the features are added back to $F_I$ to obtain $F_{IC}$.
\begin{align}
    R_{IC} &= F_I - F_C,\\
    F_{IC} &= conv(R_{IC}) + F_I.
\end{align}
%where $x$ is the input to the trident structure. $H_{is-nl}$ and $H_{cs-nl}$ are functions of non-local branch and cross-scale non-local branch, respectively. 

Intuitively, the residual feature $R_{IC}$ represents the details existing in one source while missing in the other. Such inter-residual projection allows the network to focus on only the distinct information between sources while bypassing the common knowledge, thus improves the discriminative ability of the network.

%To go a step further, $F_{res}$ is then processed by a convolution layer and added back to the target feature
%\begin{align}
%    H_{nl}(x) = F_{is-nl}(x) + H_{conv}(F_{res})
%\end{align}
%where $H_{nl}$ and $H_{conv}$ represent the combined cross/in non-local feature and convolution operator, respectively. Such source-discriminative projection allow the network to focus on distinct information between sources while bypass the common knowledge existed in both sources, thus improves the discriminative ability of the network.

%\paragraph{Back-Projection}
%While the combined feature globally exploits cross-scale and in-scale non-local similarities, it may not well fail to obtain some image properties, such as smoothness and continuity, due to the lack of local information. 

Motivated by the traditional Image SR and recent DBPN \cite{haris2018deep}, we adopt the back-projection approach to incorporate local information to regularize the feature and correct reconstruction errors. Following \cite{haris2018deep}, the final fused feature $H$ is computed by,
\begin{align}
    e &= F_L - downsample(F_{IC}), \\
    H &= upsample(e) + F_{IC},
\end{align}
where $F_L$ is the feature maps of the Local branch, $downsample$ is a stride convolution to down-sample $F_{IC}$, and $upsample$ is a stride deconvolution to upscale the feature maps.

The mutual-projected operation guarantees a residual learning while fusing different feature sources, enabling a more discriminative feature learning compared with trivial adding or concatenating.

%are first downscaled through a stride convolution to compute the reconstruction error 
%\begin{align}
%    e = x - H_{conv}(H_{nl}(x))
%\end{align}
%We then project back the error $e$ to $H_{nl}$ through a deconvolution layer
%\begin{align}
%    F_{SF} = H_{nl} + H_{deconv}(e)
%\end{align}
%where $F_{SF}$ is the final estimated SR feature from one recurrence of RTFM.

%% file: 4_3_framework.tex
\subsection{Recurrent Framework}
The repeated SEM cells are embedded into a recurrent framework, as shown in Figure \ref{fig:model}. At each iteration $i$, the hidden unit $H_i$ of SEM is directly the fused feature map $H$, and the output unit $L_{i}$ is the computed by $H_{i}$ going through a two-layer CNN. Note that the initial features $L_0$ are directly computed by the LR image $I_{LR}$ through only two convolutional layers.   

%As shown in Fig. \ref{fig:model}, our TridentSR consists of 3 parts: initial feature extraction, recurrent trident fusion module based deep feature extraction, and the reconstruction convolution. Let $I_{LR}$ and $I_{SR}$ to be the input and the output of TridentSR. As explored in [,], we apply only 2 convolution layers to extract the initial feature $F_{0}$ from the LR image.

%\begin{align}
%    F_{0} = H_{IF}(I_{LR})
%\end{align}
% where $H_{IF}$ is a composition of two convolution operators. The extracted initial feature $F_{0}$ is then used for recurrent trident fusion module (RTFM) based deep feature extraction. Considering $i$th recurrence, RTFM generates a deep LR feature $F_{LF,i}$ and a deep SR feature $F_{SF,i}$ from the previous LR feature.
 %\begin{align}
 %    &F_{LF,1}, F_{SF,1} = H_{RTFM}(F_{0}) \\
 %    &F_{LF,i}, F_{SF,i} = H_{RTFM}(F_{LF,i-1})
 %\end{align}
 %where $H_{RTFM}$ represents the RTFM based deep feature extraction module, with the deep LR feature and deep SR feature extracted in a coarse-to-fine manner. 
 
 %The proposed RTFM consists of several local convolutions, a non-local modules and a cross-scale non-local module. Therefore, the proposed RTFM is able to capture useful information in a comprehensive way. 
 
Later on, the extracted deep SR features $H_i$ from each iteration $i$ are concatenated together into a wide tensor and mapped to the SR image $I_{SR}$ via one single convolution operation. The network is trained solely with $L_1$ reconstruction loss.
%Given a training set with $N$ paired LR-HR images $(I_{LR}^{k},I_{HR}^{k})$, our goal is to optimize the L1 reconstruction

%error between $I_{HR}^k$ and $I_{SR}^k=D(I_{LR}^k;\Theta)$,
%\begin{align}
%    \mathcal{L}_{1}(\Theta) = \frac{1}{N} \sum_{k=1}^{N}\|D(I_{LR}^k;\Theta)-I_{HR}^{k}\|_1,
%\end{align}
%where $D$ represents the whole network structure, and $\Theta$ is its learnable parameters.

%% file: 5_Experiments.tex
\section{Experiments}
\subsection{Datasets and Evaluation Metrics}
Following \cite{lim2017enhanced,zhang2019residual,zhang2018residual}, we use 800 images from DIV2K \cite{timofte2017ntire} dataset to train our models. For testing,  we report the performance on five standard benchmark datasets: Set5 \cite{bevilacqua2012low}, Set14 \cite{zeyde2010single}, B100 \cite{martin2001database}, Urban100 \cite{huang2015single} and Manga109 \cite{matsui2017sketch}. For evaluation, all the SR results are first transformed into YCbCr space and evaluated by PSNR and SSIM \cite{wang2004image} metrics
on Y channel only.

\subsection{Implementation details}
We set the recurrence number of SEM as 12 following \cite{liu2018non}. For the Cross-Scale Non-Local (CS-NL) attention in SEM, we set patch size $p=3$ and stride $s=1$ for dense sampling. We use $3\times 3$ as filter size for all convolution layers except for those in attention blocks where the kernel size is $1\times 1$. The filter size for stride convolution and deconvolution in SEM are set to be equal at each scale, e.g., $6\times 6$, $9\times 9$ and $8\times8$ for scale factor 2, 3, 4, respectively. All intermediate features have channel $C=128$ except for those embedded features in attention module, where $C=64$. The last convolution layer in SEM has 3 convolution filters that transfer a deep SR feature to an RGB image. 

During training, we crop 16 images with patch size $48\times 48$ to form a input batch. The training examples are augmented by random rotating \ang{90}, \ang{180}, \ang{270} and horizontal flipping. To optimize our model, we use ADAM optimizer \cite{kingma2014adam} with $\beta_1=0.9$, $\beta_2=0.999$, and $\epsilon=$1e-8. The initial learning rate is set to 1e-4 and reduced to half every 150 epochs. The training stops at 500 epochs. We implement the model using PyTorch, and train it on Nvidia V100 GPUs. 

\subsection{Comparisons with State-of-the-arts}
\input{./experiment/main_table.tex}

\input{./experiment/main_figure.tex}
To verify the effectiveness of the proposed model, we compare our approach with 11 state-of-the-art methods, which are LapSRN \cite{lai2017deep}, SRMDNF \cite{zhang2018learning}, MemNet \cite{tai2017memnet}, EDSR \cite{lim2017enhanced}, DBPN \cite{haris2018deep}, RDN \cite{zhang2018residual}, RCAN \cite{zhang2018image}, NLRN\cite{liu2018non}, SRFBN \cite{li2019feedback}, OISR \cite{he2019ode} and SAN \cite{dai2019second}. %Similar to \qa{[cite]}, we also adopt self-ensemble approach to further improve our TridentSR and denote it as TridentSR+. 

\paragraph{Quantitative Evaluations}
In Table \ref{tab:results_psnr_ssim_x2348}, We report the quantitative comparisons for scale factor $\times$2, $\times$3 and $\times$4. Compared with other methods, our CS-NL-embedded recurrent model achieved the best performance on multiple benchmarks for almost all scaling factors. %Even without self-ensemble, our methods still outperforms most of other methods. 
It worth noting that our model significantly outperforms NLRN, which is the first proposed in-scale non-local network for image restoration. 

Our method has better performance when the scaling factor is larger. For $\times 4$ settings, our CS-NL embedded model achieves the state-of-the-art PSNR for all the testing benchmarks. In particular, for Urban100 and Manga109 dataset, our model outperforms previous state-of-the-art approaches by 0.4 dB and 0.2 dB, respectively. These datasets contains abundant repeated patterns, such as edges and small corners. Therefore, the superior performance demonstrates the effectiveness of our attention in exploiting internal HR hints. We claim that cross-scale intrinsic priors are indeed effective for a more faithful reconstruction.
\paragraph{Qualitative Evaluations}
The qualitative evaluations on Urban100 dataset are shown in Figure \ref{tab:table 4}. The proposed model is proven to be more effective for images with repeated high-frequency features like windows, lines, squares, etc. For example, in the figure of building, LR image contains plenty of window features covering long-range of spatial-frequency. Directly utilizing those cross-scale self-exemplars from the images will be significantly better than searching for in-scale features or external patches in the training set. For all the shown examples, our method perceptually out-performs other state-of-the-arts by a large margin.
\begin{table}[h]
\footnotesize
%\small
    \centering
    \begin{tabular}{|c|c|c|c|c|c|c|}
    \hline
        &EDSR & DBPN & RDN & RCAN & SAN & CSNLN \\
         \hline
         Para. & 43M & \color{blue}{10M} & 22.3M & 16M & 15.7M & \color{red}{3M}\\
         \hline
         PSNR & 38.11 & 38.09 & 38.24 & 38.27 & \color{red}{38.31} &\color{blue}{38.28}
         \\ \hline
    \end{tabular}
    \caption{Model size and performance comparsion on Set5 (2$\times$) .}
    \label{tab:table model}
\end{table}

\noindent\textbf{Model Size Analysis}
We report the model size and performance for recently competitive SR methods in Table \ref{tab:table model}. Comparing with others, our model has the least parameters, which only needs 20\% parameters of RCAN and SAN, but achieves the second best result. Therefore, our CSNLN obtains better parameter efficiency in comparison with other methods, by effectively mining internal HR hints. 

\subsection{Ablation Study}
\begin{figure*}[t]
 \centering
 \begin{minipage}{0.105\textwidth}
    \includegraphics[ width = 1\textwidth]{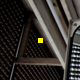}
 \end{minipage}
 \begin{minipage}{0.105\textwidth}
    \includegraphics[ width = 1\textwidth]{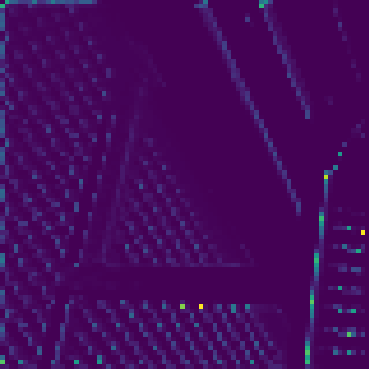}
 \end{minipage}
 \begin{minipage}{0.105\textwidth}
    \includegraphics[ width = 1\textwidth]{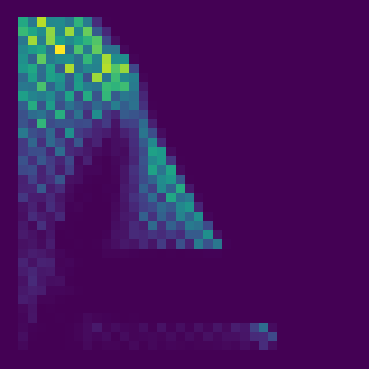}
 \end{minipage}
 \begin{minipage}{0.105\textwidth}
    \includegraphics[ width = 1\textwidth]{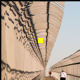}
 \end{minipage}
 \begin{minipage}{0.105\textwidth}
    \includegraphics[ width = 1\textwidth]{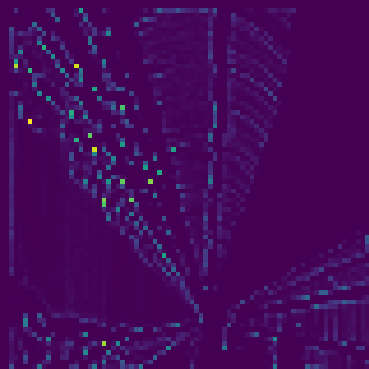}
 \end{minipage}
 \begin{minipage}{0.105\textwidth}
    \includegraphics[ width = 1\textwidth]{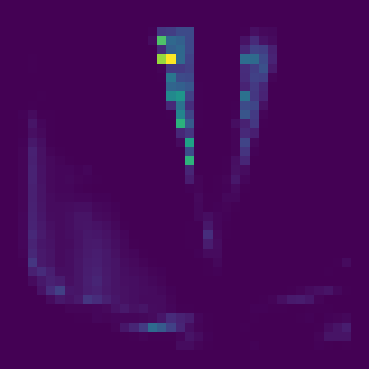}
 \end{minipage}
 \begin{minipage}{0.105\textwidth}
    \includegraphics[ width = 1\textwidth]{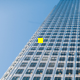}
 \end{minipage}
 \begin{minipage}{0.105\textwidth}
    \includegraphics[width = 1\textwidth]{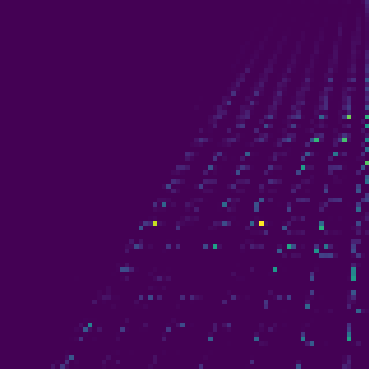}
 \end{minipage}
 \begin{minipage}{0.105\textwidth}
    \includegraphics[ width = 1\textwidth]{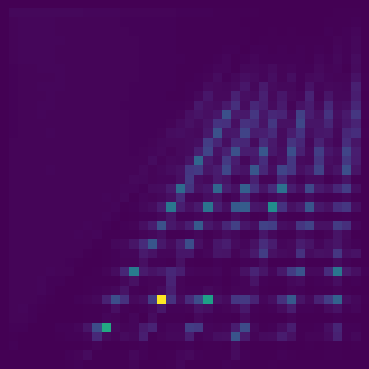}
 \end{minipage}
 \begin{minipage}{0.105\textwidth}
    \includegraphics[ width = 1\textwidth]{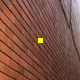}
 \end{minipage}
 \begin{minipage}{0.105\textwidth}
    \includegraphics[ width = 1\textwidth]{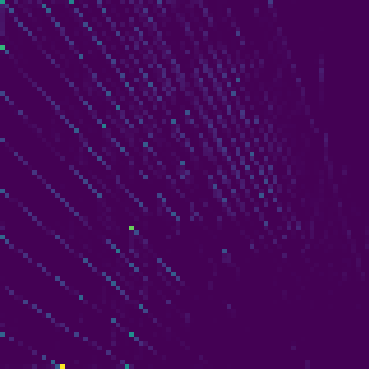}
 \end{minipage}
 \begin{minipage}{0.105\textwidth}
    \includegraphics[ width = 1\textwidth]{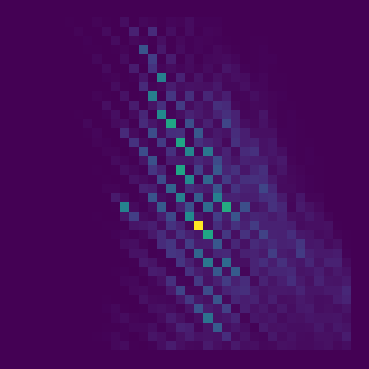}
 \end{minipage}
 \begin{minipage}{0.105\textwidth}
    \includegraphics[ width = 1\textwidth]{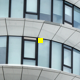}
 \end{minipage}
 \begin{minipage}{0.105\textwidth}
    \includegraphics[width = 1\textwidth]{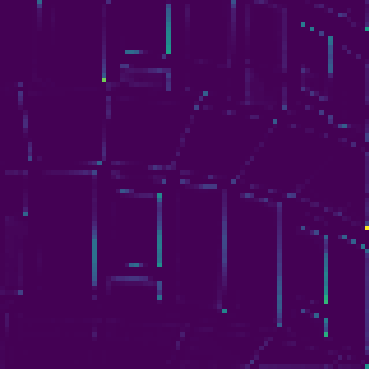}
 \end{minipage}
 \begin{minipage}{0.105\textwidth}
    \includegraphics[ width = 1\textwidth]{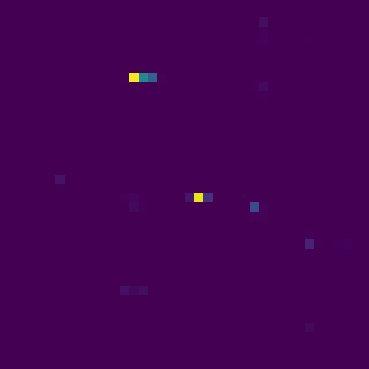}
 \end{minipage}
 \begin{minipage}{0.105\textwidth}
    \includegraphics[ width = 1\textwidth]{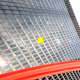}
 \end{minipage}
 \begin{minipage}{0.105\textwidth}
    \includegraphics[ width = 1\textwidth]{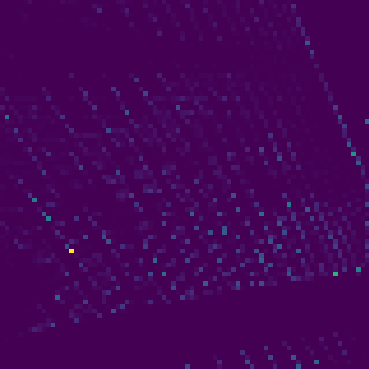}
 \end{minipage}
 \begin{minipage}{0.105\textwidth}
    \includegraphics[ width = 1\textwidth]{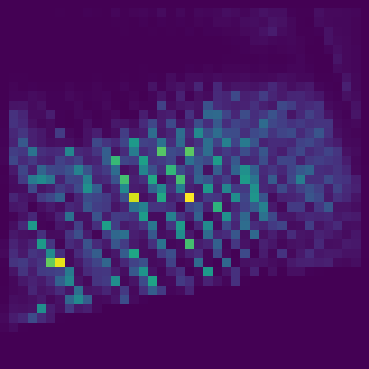}
 \end{minipage}
 \\
 \centering
 \vspace{1.5mm}
\caption{Comparisons of correlation maps of CS-NL attention and IS-NL attention. For each group of three columns, the left one is the input image, the middle one shows the in-scale attention, and the right one depicts the cross-scale attention. one can see that the in-scale attention only focuses on pixels with similar intensity. In contrast, our cross-scale non-local attention is able to utilize the abundant repeated structures in the images, demonstrating its effectiveness for exploiting internal HR information. }
\label{fig:att}
\end{figure*}

\begin{table*}[]\setlength{\tabcolsep}{8pt}
%\footnotesize
%\small
    \centering
    \resizebox{\linewidth}{!}{
    \begin{tabular}{|l|c|c|c|c|c|c|c|c|}
    \hline
    Local (L) branch &  & \cmark &&  &\cmark &\cmark & &\cmark  \\
    \cline{1-1}
         In-Scale Non-Local (IS-NL) Branch & & &\cmark &  &\cmark &  & \cmark & \cmark
         \\
         \cline{1-1}
         Cross-Scale Non-Local (CS-NL) branch&  &  & &\cmark  & &\cmark &\cmark & \cmark \\
         \hline
         PSNR & 33.32 &33.47 &33.52 &33.51 &33.62 &33.64 &33.57 &\textbf{33.74}\\
         \hline
    \end{tabular}}
    %\vspace{0.5mm}
    \caption{Ablation study on the branch features in SEM. We report the PSNR results on Set14 (2$\times$) after 200 epochs. With an additional CS-NL branch, the performance becomes 33.74dB compared with the one without CS-NL, 33.62dB.}
    \vspace{-1mm}
    \label{tab:table 1}
\end{table*}

\begin{table}[]
%\footnotesize
%\small
    \centering
    \resizebox{\linewidth}{!}{
    \begin{tabularx}{\linewidth}{|l|Y|Y|Y|}
    \hline
         Attention Patch Size & 1$\times$1 & 3$\times$3 & 5$\times$5 \\
         \hline
         PSNR & 33.67 &\textbf{33.74} &33.61\\
         \hline
    \end{tabularx}}
    %\vspace{0.2mm}
    \caption{Effects of patch size for matching.}
    \vspace{-1mm}
    \label{tab:table 2}
\end{table}

\begin{table}[]
%\footnotesize
%\small
    \centering
    \resizebox{\linewidth}{!}{
    \begin{tabularx}{\linewidth}{|l|Y|Y|Y|}
    \hline
         Fusion & addition & concatenation & Mutual Projection  \\
         \hline
         PSNR &33.69 &33.62 &\textbf{33.74} \\
         \hline
    \end{tabularx}}
    %\vspace{0.2mm}
    \caption{Comparison of fusion operators.}
    \label{tab:table 3}
\end{table}

\paragraph{Cross-Scale v.s. In-Scale Attention}
The key difference between our cross-scale non-local attention and the in-scale one is to allow network to benefit from abundant internal HR hints with different scales. To verify it, we visualize its correlation maps on 6 images from Urban100 \cite{huang2015single}, and compare it with in-scale non-local attention. 

As shown in Figure \ref{fig:att}, these images contain extensive recurrences of small patterns both within scale and across scale. It is interesting to point out that once the image contains repeated edges, such redundant recurrences are not limited to where high scale patterns appear, but also can be found in-place or even in the area that pattern tends to slightly shrink. For example, the HR appearance of a small corner can be simply found by properly zooming out. All these recurrences contain valuable high frequency information for reconstruction. As shown in Figure \ref{fig:att}, the in-scale attention only focuses on pixels with similar intensity. In contrast, our cross-scale non-local attention is able to utilize the abundant repeating structures in the images, demonstrating its effectiveness for exploiting internal HR information. 

\paragraph{Self-Exemplars Mining Module} 
To demonstrate the effectiveness of our proposed Self-Exemplars Mining (SEM) module, we construct a baseline model by removing all branches, resulting in a fully convolutional recurrent network (RNN). To keep the total parameters same as other variants, we set 10 convolution layers in the recurrent block. As shown in Table \ref{tab:table 1}, the basic RNN achieves 33.32 dB on Set14 ($\times2$). Results in first 4 columns demonstrate the effectiveness of individual branch, as each of them brings improvement over the baseline. Furthermore, from last 4 columns, we find that combining these branches achieves the best performance. For example, when cross-scale non-local branch is added, the performance is improved from 33.47 dB to 33.64 dB. When both local branch and non-local branch are added to the network, the best performance is achieved by further adding cross-scale non-local branch, resulting in a improvement from 33.62 dB to 33.74 dB. 

These facts indicate that the cross-scale correlations learned by our attention can not be captured by either simple convolution or previous in-scale attention module, demonstrating that our CS-NL attention is of crucial importance for fully exploiting information from LR images.

\paragraph{Patch-Based Matching v.s. Pixel-Based Matching} In practical implementation, we compute patch-wise correlation rather than pixel-wise correlation. Here we investigate the influence of patch size $p$ in CS-NL attention. We compare the patch size of $1\times 1$, $3\times 3$ and $5\times 5$, where $1\times 1$ is equivalent to pixel-wise matching. As shown in Table \ref{tab:table 2}, the performance peak is at $3\times 3$, which is higher than pixel based matching, indicating that a small patch can serve as a better region descriptor. However, when using a larger patch size, the performance is worse than the pixel-based matching. This is mainly because larger patches mean additional constraint on the content when evaluating similarity, and therefore it becomes harder to find well-matched correspondences. All these results show that it is necessary to choose a proper patch size for effectively computing correlations in CS-NL attention.

\paragraph{Mutual-Projected Fusion} We show the effectiveness of our mutual-projected fusion by comparing it with other commonly used fusion strategies, e.g., feature addition and concatenation. As shown in Table \ref{tab:table 3}, it can be found that our projection based fusion obtains the best result. By replacing the addition and concatenation with mutual projection, the performance improves about 0.05 dB and 0.12 dB. These results demonstrate the effectiveness of our fusion module in progressively aggregating information. 

%\textbf{Network depth.}

%% file: experiment/main_table.tex
\begin{table*}[thbp]\setlength{\tabcolsep}{7pt}
%\scriptsize
%\footnotesize
\small
%\normalsize
\center
\begin{center}
\caption{Quantitative results on benchmark datasets. Best and second best results are colored with \textcolor{red}{red} and \textcolor{blue}{blue}.}
\label{tab:results_psnr_ssim_x2348}
\begin{tabular}{|l|c|c|c|c|c|c|c|c|c|c|c|}
\hline
\multirow{2}{*}{Method} & \multirow{2}{*}{Scale} &  \multicolumn{2}{c|}{Set5} &  \multicolumn{2}{c|}{Set14} &  \multicolumn{2}{c|}{B100} &  \multicolumn{2}{c|}{Urban100} &  \multicolumn{2}{c|}{Manga109}  
\\
%\hline
\cline{3-12}
&  & PSNR & SSIM & PSNR & SSIM & PSNR & SSIM & PSNR & SSIM & PSNR & SSIM 
\\
\hline
\hline

LapSRN~\cite{lai2017deep} & $\times$2 
& 37.52
 & 0.9591
  & 33.08
   & 0.9130
    & 31.08
     & 0.8950
      & 30.41
       & 0.9101
        & 37.27
         & 0.9740
                   
\\
MemNet~\cite{tai2017memnet} & $\times$2 
& 37.78
 & 0.9597
  & 33.28
   & 0.9142
    & 32.08
     & 0.8978
      & 31.31
       & 0.9195
        & 37.72
         & 0.9740
                   
\\
EDSR~\cite{lim2017enhanced} & $\times$2 
& 38.11
 & 0.9602
  & 33.92
   & 0.9195
    & 32.32
     & 0.9013
      & 32.93
       & 0.9351
        & 39.10
         & 0.9773
                   
\\
SRMDNF~\cite{zhang2018learning} & $\times$2 
& 37.79
 & 0.9601
  & 33.32
   & 0.9159
    & 32.05
     & 0.8985
      & 31.33
       & 0.9204
        & 38.07
         & 0.9761
                   
\\
DBPN~\cite{haris2018deep} & $\times$2 
& 38.09
 & 0.9600
  & 33.85
   & 0.9190
    & 32.27
     & 0.9000
      & 32.55
       & 0.9324
        & 38.89
         & 0.9775        
\\
RDN~\cite{zhang2018residual} & $\times$2 
& 38.24
 & 0.9614
  & 34.01
   & 0.9212
    & 32.34
     & 0.9017
      & 32.89
       & 0.9353
        & 39.18
         & 0.9780
         
\\

RCAN~\cite{zhang2018image} & $\times$2 
& {38.27}
 & {0.9614}
  & \color{red}{34.12}
   & \color{blue}{0.9216}
    & \color{blue}{32.41}
     & \color{blue}{0.9027}
      & \color{red}{33.34}
       & \color{blue}{0.9384}
        & \color{red}{39.44}
         & \color{blue}{0.9786}
\\       
NLRN~\cite{liu2018non}& $\times$2 
& {38.00}
 & {0.9603}
  & {33.46}
   & {0.9159}
    & {32.19}
     & {0.8992}
      & {31.81}
       & {0.9249}
        & {--}
         & {--}
\\
SRFBN~\cite{li2019feedback}& $\times$2 
& {38.11}
 & {0.9609}
  & {33.82}
   & {0.9196}
    & {32.29}
     & {0.9010}
      & {32.62}
       & {0.9328}
        & {39.08}
         &{0.9779}
\\
 OISR~\cite{he2019ode} & $\times$2 
& {38.21}
 & {0.9612}
  & {33.94}
   & {0.9206}
    & {32.36}
     & {0.9019}
      & {33.03}
       & {0.9365}
        &--
         & --      
\\
SAN~\cite{dai2019second} & $\times$2 
& \color{red}{38.31}
 & \color{red}{0.9620}
  & \color{blue}{34.07}
   & {0.9213}
    & \color{red}{32.42}
     & \color{red}{0.9028}
      & {33.10}
       & {0.9370}
        & {39.32}
         & \color{red}{0.9792}
\\
CSNLN (ours) & $\times$2 
& \color{blue}{38.28}
 & \color{blue}{0.9616}
  & \color{red}{34.12}
   & \color{red}{0.9223}
    & {32.40}
     & {0.9024}
      & \color{blue}{33.25}
       & \color{red}{0.9386}
        & \color{blue}{39.37}
         & {0.9785}
\\

\hline
\hline

LapSRN~\cite{lai2017deep} & $\times$3 
& 33.82
 & 0.9227
  & 29.87
   & 0.8320
    & 28.82
     & 0.7980
      & 27.07
       & 0.8280
        & 32.21
         & 0.9350
                   
\\
MemNet~\cite{tai2017memnet} & $\times$3 
& 34.09
 & 0.9248
  & 30.00
   & 0.8350
    & 28.96
     & 0.8001
      & 27.56
       & 0.8376
        & 32.51
         & 0.9369
                   
\\
EDSR~\cite{lim2017enhanced} & $\times$3 
& 34.65
 & 0.9280
  & 30.52
   & 0.8462
    & 29.25
     & 0.8093
      & 28.80
       & 0.8653
        & 34.17
         & 0.9476
                   
\\
SRMDNF~\cite{zhang2018learning} & $\times$3 
& 34.12
 & 0.9254
  & 30.04
   & 0.8382
    & 28.97
     & 0.8025
      & 27.57
       & 0.8398
        & 33.00
         & 0.9403
                   
\\
RDN~\cite{zhang2018residual} & $\times$3 
& 34.71
 & 0.9296
  & 30.57
   & 0.8468
    & 29.26
     & 0.8093
      & 28.80
       & 0.8653
        & 34.13
         & 0.9484
         
\\
RCAN~\cite{zhang2018image}& $\times$3 
& \color{blue}{34.74}
 &\color{blue}{0.9299}
  & \color{blue}{30.65}
   & \color{red}{0.8482}
    & \color{blue}{29.32}
     & \color{blue}{0.8111}
      & \color{blue}{29.09}
       &\color{blue}{0.8702}
        & \color{blue}{34.44}
         &\color{blue}{0.9499}
         
\\
NLRN~\cite{liu2018non}& $\times$3 
& {34.27}
 &{0.9266}
  & {30.16}
   &{0.8374}
    & {29.06}
     & {0.8026}
      & {27.93}
       & {0.8453}
        & {-}
         & {-}
\\
SRFBN~\cite{li2019feedback}& $\times$3 
& {34.70}
 &{0.9292}
  & {30.51}
   &{0.8461}
    & {29.24}
     & {0.8084}
      & {28.73}
       & {0.8641}
        & {34.18}
         & {0.9481}
\\
OISR~\cite{he2019ode}& $\times$3 
& {34.72}
 &{0.9297}
  & {30.57}
   &{0.8470}
    & {29.29}
     & {0.8103}
      & {28.95}
       & {0.8680}
        & {-}
         & {-}
\\
SAN~\cite{dai2019second} & $\times$3 
& \color{red}{34.75}
 &\color{red}{0.9300}
  & {30.59}
   &\color{blue}{0.8476}
    &\color{red}{29.33}
     & \color{red}{0.8112}
      & {28.93}
       & {0.8671}
        & {34.30}
         & {0.9494}
        
\\
CSNLN (ours) & $\times$3
& \color{blue}{34.74}
 & \color{red}{0.9300}
  & \color{red}{30.66}
   & \color{red}{0.8482}
    & \color{red}{29.33}
     & {0.8105}
      & \color{red}{29.13}
       & \color{red}{0.8712}
        & \color{red}{34.45}
         & \color{red}{0.9502}
\\
\hline
\hline
LapSRN~\cite{lai2017deep} & $\times$4 
& 31.54
 & 0.8850
  & 28.19
   & 0.7720
    & 27.32
     & 0.7270
      & 25.21
       & 0.7560
        & 29.09
         & 0.8900
                   
\\
MemNet~\cite{tai2017memnet} & $\times$4 
& 31.74
 & 0.8893
  & 28.26
   & 0.7723
    & 27.40
     & 0.7281
      & 25.50
       & 0.7630
        & 29.42
         & 0.8942
                   
\\
EDSR~\cite{lim2017enhanced} & $\times$4 
& 32.46
 & 0.8968
  & 28.80
   & 0.7876
    & 27.71
     & 0.7420
      & 26.64
       & 0.8033
        & 31.02
         & 0.9148
                   
\\
SRMDNF~\cite{zhang2018learning} & $\times$4 
& 31.96
 & 0.8925
  & 28.35
   & 0.7787
    & 27.49
     & 0.7337
      & 25.68
       & 0.7731
        & 30.09
         & 0.9024
                   
\\
DBPN~\cite{haris2018deep} & $\times$4 
& 32.47
 & 0.8980
  & 28.82
   & 0.7860
    & 27.72
     & 0.7400
      & 26.38
       & 0.7946
        & 30.91
         & 0.9137
         
\\
RDN~\cite{zhang2018residual} & $\times$4 
& 32.47
 & 0.8990
  & 28.81
   & 0.7871
    & 27.72
     & 0.7419
      & 26.61
       & 0.8028
        & 31.00
         & 0.9151
         
\\
RCAN~\cite{zhang2018image}& $\times$4 
& {32.63}
 & {0.9002}
  & {28.87}
   &\color{red}{0.7889}
    & {27.77}
     & \color{blue}{0.7436}
      &\color{blue} {26.82}
       & \color{blue}{0.8087}
        &\color{blue}{31.22}
         & \color{blue}{0.9173}

\\
NLRN~\cite{liu2018non}& $\times$4 
& {31.92}
 & {0.8916}
  & {28.36}
   & {0.7745}
    & {27.48}
     & {0.7306}
      & {25.79}
       & {0.7729}
        & {-}
         & {-}
\\
SRFBN~\cite{li2019feedback} & $\times$4 
& {32.47}
 & {0.8983}
  & {28.81}
   & {0.7868}
    & {27.72}
     & {0.7409}
      & {26.60}
       & {0.8015}
        & {31.15}
         & {0.9160}
\\
OISR~\cite{he2019ode} & $\times$4 
&{32.53}
 &{0.8992}
  &{28.86}
   & {0.7878}
    &{27.75}
     & {0.7428}
      & {26.79}
       & {0.8068}
        & {-}
         & {-}
\\
SAN~\cite{dai2019second} & $\times$4 
& \color{blue}{32.64}
 &\color{blue}{0.9003}
  &\color{blue}{28.92}
   &\color{blue}{0.7888}
    &\color{blue}{27.78}
     & \color{blue}{0.7436}
      & {26.79}
       & {0.8068}
        & {31.18}
         & {0.9169}
\\
CSNLN (ours)  & $\times$4 
& \color{red}{32.68}
 & \color{red}{0.9004}
  & \color{red}{28.95}
   & \color{blue}{0.7888}
    & \color{red}{27.80}
     & \color{red}{0.7439}
      & \color{red}{27.22}
       & \color{red}{0.8168}
        & \color{red}{31.43}
         & \color{red}{0.9201}

\\
\hline             
\end{tabular}
\end{center}
\end{table*}

%% file: experiment/main_figure.tex
\begin{figure*}[htbp]
%\tiny
%\small
	\newlength\fsdttwofigBD
	\setlength{\fsdttwofigBD}{-5.0mm}
	\scriptsize
	\centering
	\begin{tabular}{cc}
	%\tiny
	%\scriptsize
	%\footnotesize
	%\small
		%\hspace{-0.4cm}
		\begin{adjustbox}{valign=t}
		%\tiny
			\begin{tabular}{c}
				\includegraphics[width=0.229\textwidth]{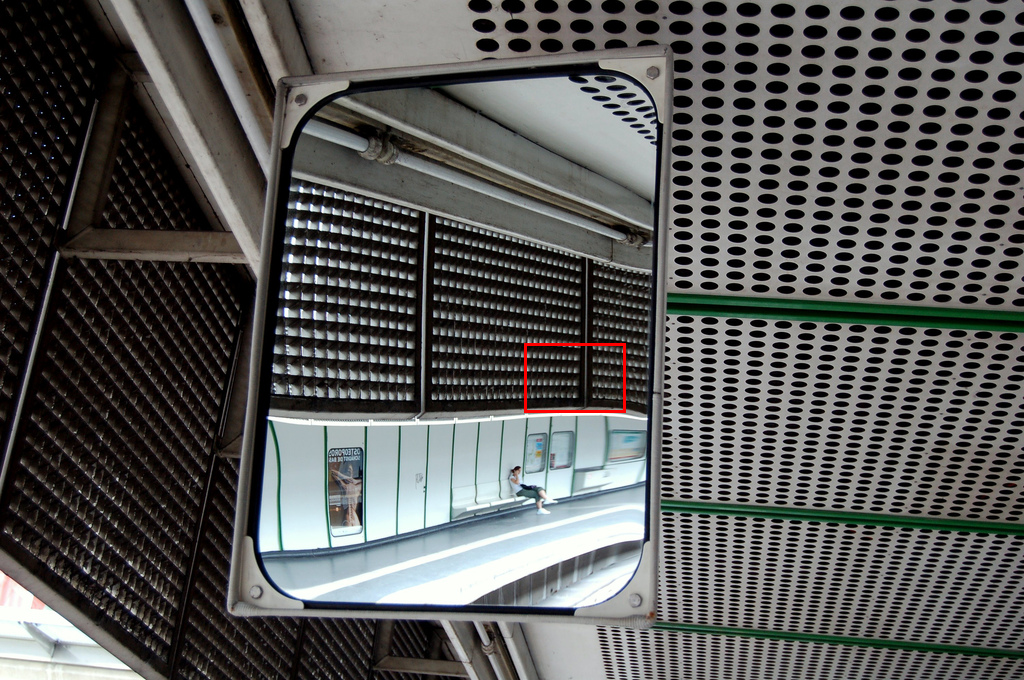}
				\\
				 Urban100 ($4\times$):
				\\
				img\_005
				%\textsc{B100}: img_092
				
			\end{tabular}
		\end{adjustbox}
		\hspace{-2.3mm}
		\begin{adjustbox}{valign=t}
		%\tiny
			\begin{tabular}{cccccc}
				\includegraphics[width=\widthscalefive \textwidth]{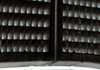} \hspace{\fsdttwofigBD} &
				\includegraphics[width=\widthscalefive \textwidth]{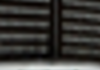} \hspace{\fsdttwofigBD} &
				\includegraphics[width=\widthscalefive \textwidth]{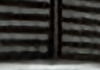} \hspace{\fsdttwofigBD} &
				\includegraphics[width=\widthscalefive \textwidth]{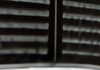} \hspace{\fsdttwofigBD} &
				\includegraphics[width=\widthscalefive \textwidth]{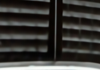} 
				\\
				HR \hspace{\fsdttwofigBD} &
				Bicubic \hspace{\fsdttwofigBD} &
				LapSRN~\cite{lai2017deep} \hspace{\fsdttwofigBD} &
				EDSR~\cite{lim2017enhanced} \hspace{\fsdttwofigBD} &
				DBPN~\cite{haris2018deep}
				\\
				\includegraphics[width=\widthscalefive \textwidth]{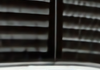} \hspace{\fsdttwofigBD} &
				\includegraphics[width=\widthscalefive \textwidth]{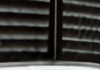} \hspace{\fsdttwofigBD} &
				\includegraphics[width=\widthscalefive \textwidth]{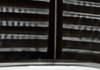} \hspace{\fsdttwofigBD} &
				\includegraphics[width=\widthscalefive \textwidth]{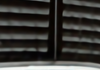} \hspace{\fsdttwofigBD} &
				\includegraphics[width=\widthscalefive \textwidth]{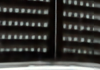}  
				\\ 
				OISR~\cite{he2019ode} \hspace{\fsdttwofigBD} &
				RDN~\cite{zhang2018residual} \hspace{\fsdttwofigBD} &
				RCAN~\cite{zhang2018image} \hspace{\fsdttwofigBD} &
				SAN~\cite{dai2019second}  \hspace{\fsdttwofigBD} &
				Ours 
			 \hspace{\fsdttwofigBD} 
				\\
			\end{tabular}
		\end{adjustbox}
		\vspace{0.5mm}
		\\

		\begin{adjustbox}{valign=t}
		%\tiny
			\begin{tabular}{c}
				\includegraphics[width=0.229\textwidth]{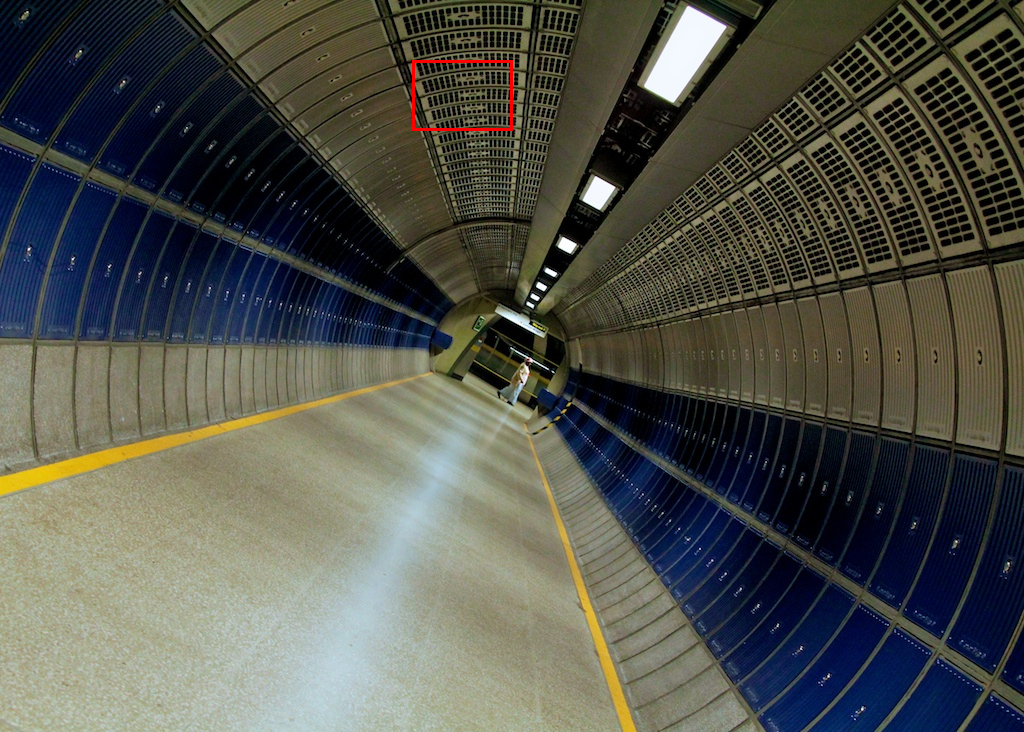}
				\\
				 Urban100 ($4\times$):
				\\
				img\_078
				%\textsc{B100}: img_092
				
			\end{tabular}
		\end{adjustbox}
		\hspace{-2.3mm}
		\begin{adjustbox}{valign=t}
		%\tiny
			\begin{tabular}{cccccc}
				\includegraphics[width=\widthscalefive \textwidth]{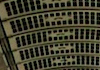} \hspace{\fsdttwofigBD} &
				\includegraphics[width=\widthscalefive \textwidth]{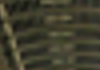} \hspace{\fsdttwofigBD} &
				\includegraphics[width=\widthscalefive \textwidth]{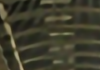} \hspace{\fsdttwofigBD} &
				\includegraphics[width=\widthscalefive \textwidth]{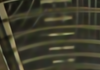} \hspace{\fsdttwofigBD} &
				\includegraphics[width=\widthscalefive \textwidth]{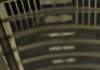} 
				\\
				HR \hspace{\fsdttwofigBD} &
				Bicubic \hspace{\fsdttwofigBD} &
				LapSRN~\cite{lai2017deep} \hspace{\fsdttwofigBD} &
				EDSR~\cite{lim2017enhanced} \hspace{\fsdttwofigBD} &
				DBPN~\cite{haris2018deep}
				\\
				\includegraphics[width=\widthscalefive \textwidth]{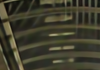} \hspace{\fsdttwofigBD} &
				\includegraphics[width=\widthscalefive \textwidth]{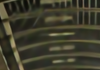} \hspace{\fsdttwofigBD} &
				\includegraphics[width=\widthscalefive \textwidth]{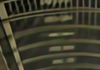} \hspace{\fsdttwofigBD} &
				\includegraphics[width=\widthscalefive \textwidth]{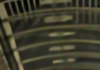} \hspace{\fsdttwofigBD} &
				\includegraphics[width=\widthscalefive \textwidth]{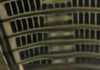}  
				\\ 
				OISR~\cite{he2019ode} \hspace{\fsdttwofigBD} &
				RDN~\cite{zhang2018residual} \hspace{\fsdttwofigBD} &
				RCAN~\cite{zhang2018image} \hspace{\fsdttwofigBD} &
				SAN~\cite{dai2019second}  \hspace{\fsdttwofigBD} &
				Ours 
			 \hspace{\fsdttwofigBD} 
				\\
			\end{tabular}
		\end{adjustbox}
		\vspace{0.5mm}
		\\
		%\hspace{-0.4cm}		
		\begin{adjustbox}{valign=t}
		%\tiny
			\begin{tabular}{c}
				\includegraphics[width=0.229\textwidth]{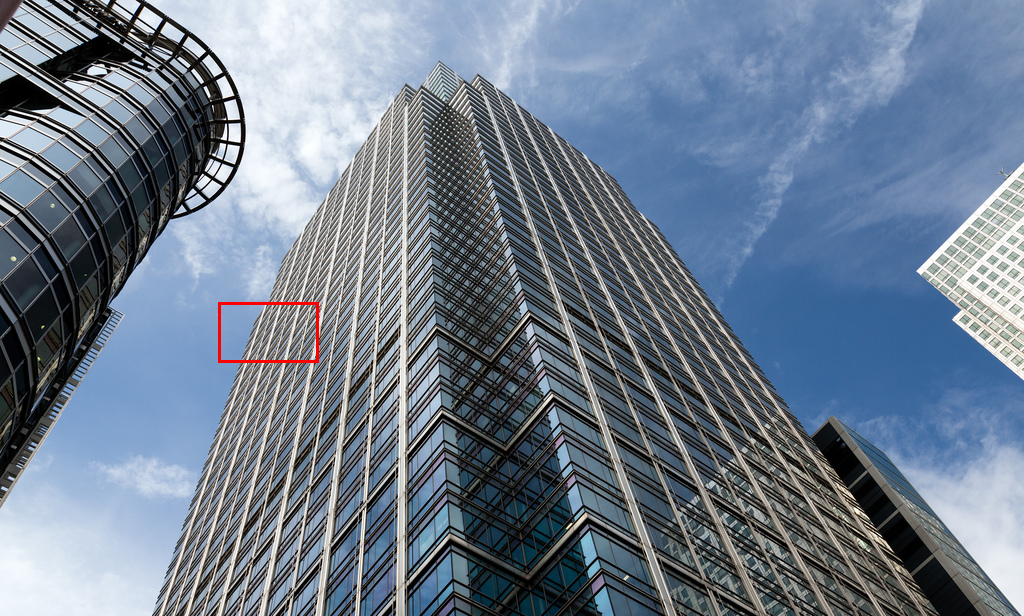}
				\\
				Urban100 ($4\times$):
				\\
				%\textsc{Urban100}: img\_004
				img\_047
			\end{tabular}
		\end{adjustbox}
		\hspace{-2.3mm}
		\begin{adjustbox}{valign=t}
		%\tiny
			\begin{tabular}{cccccc}
				\includegraphics[width=\widthscalefive \textwidth]{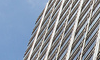} \hspace{\fsdttwofigBD} &
				\includegraphics[width=\widthscalefive \textwidth]{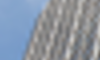} \hspace{\fsdttwofigBD} &
				\includegraphics[width=\widthscalefive \textwidth]{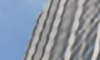} \hspace{\fsdttwofigBD} &
				\includegraphics[width=\widthscalefive \textwidth]{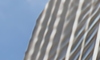} \hspace{\fsdttwofigBD} &
				\includegraphics[width=\widthscalefive \textwidth]{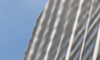} 
				\\
				HR \hspace{\fsdttwofigBD} &
				Bicubic \hspace{\fsdttwofigBD} &
				LapSRN~\cite{lai2017deep} \hspace{\fsdttwofigBD} &
				EDSR~\cite{lim2017enhanced} \hspace{\fsdttwofigBD} &
				DBPN~\cite{haris2018deep}
				\\
				\includegraphics[width=\widthscalefive \textwidth]{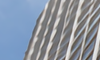} \hspace{\fsdttwofigBD} &
				\includegraphics[width=\widthscalefive \textwidth]{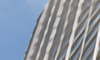} \hspace{\fsdttwofigBD} &
				\includegraphics[width=\widthscalefive \textwidth]{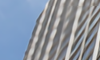} \hspace{\fsdttwofigBD} &
				\includegraphics[width=\widthscalefive \textwidth]{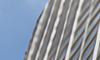} \hspace{\fsdttwofigBD} &
				\includegraphics[width=\widthscalefive \textwidth]{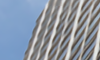}  
				\\ 
				OISR~\cite{he2019ode} \hspace{\fsdttwofigBD} &
				RDN~\cite{zhang2018residual} \hspace{\fsdttwofigBD} &
				RCAN~\cite{zhang2018image} \hspace{\fsdttwofigBD} &
				SAN~\cite{dai2019second}  \hspace{\fsdttwofigBD} &
				Ours
				\\
				
				\\
			\end{tabular}
		\end{adjustbox}
	\end{tabular}
	\caption{
		Visual comparison for $4\times$ SR on Urban100 dataset. For all the shown examples, especially the images with repeated edges or structures, our method perceptually out-performs other state-of-the-arts by a large margin.
	}
	\label{tab:table 4}
\vspace{-3mm}
\end{figure*}

%% file: 6_Conclusion.tex
\section{Conclusion}
In this paper, we proposed the first Cross-Scale Non-Local (CS-NL) attention module for image super-resolution deep networks. With the novel module, we are able to sufficiently discover the widely existing cross-scale feature similarities in natural images. Further integrating it with local and the previous in-scale non-local priors, while embracing the abundant external information learned by the network, our recurrent model achieved state-of-the-art performance for multiple benchmarks. Our experiments suggest that exploring cross-scale long-range dependencies will greatly benefit single image super-resolution (SISR) task, and possibly is also promising for general image restoration task. 